\documentclass[acmtog,nonacm]{acmart}

\setcopyright{none}
\renewcommand\footnotetextcopyrightpermission[1]{}
\AtBeginDocument{%
  }

\definecolor{projectgreen}{HTML}{0B8A3A}

\usepackage{subcaption}
\begin{document}

\title{LUCID: Learning Unified Control for Image Deflaring and Exposure Mastery in Nighttime Photography}

\author{Tingyu Yang}
\orcid{0009-0004-5842-8680}
\affiliation{%
  \department{MoE Key Lab of Artificial Intelligence, AI Institute, School of Computer Science; School of Biomedical Engineering}
  \institution{Shanghai Jiao Tong University}
  \city{Shanghai}
  \country{China}}
\email{frakenation@sjtu.edu.cn}

\author{Yuan Cheng}
\orcid{0000-0001-9806-3613}
\affiliation{%
\department{School of Artificial Intelligence}
\institution{Shanghai Jiao Tong University}
\city{Shanghai}
\country{China}}
\email{cyuan328@sjtu.edu.cn}

\author{Xiaoyun Yuan}
\authornote{Corresponding author.}
\orcid{0000-0002-7914-3658}
\affiliation{%
\department{MoE Key Lab of Artificial Intelligence, AI Institute, School of Computer Science}
\institution{Shanghai Jiao Tong University}
\city{Shanghai}
\country{China}}
\email{yuanxiaoyun@sjtu.edu.cn}

\begin{abstract}
Photography is the art of painting with light, yet nighttime scenes are shaped by competing degradations: intense flares obscure scene structure, while photon-limited regions collapse into noise. Conventional approaches address these factors in isolation, overlooking the fact that these degradations are fundamentally entangled. To bridge this gap, we introduce LUCID, a unified framework that reframes nighttime restoration as a continuous and controllable process rather than a fixed correction.
We decompose nighttime restoration into two cooperative components: a flare disentanglement module that lifts the 'curtain' of optical artifacts to provide reliable structural guidance, and a diffusion-driven module that leverages generative priors to reconstruct clean and well-exposed imagery.
Crucially, LUCID introduces explicit controllability through a novel four-mode training strategy, enabling users to steer the restoration process via classifier-free guidance (CFG) and allowing selective control over light sources and their associated flare and ghosting artifacts, while also supporting high dynamic range (HDR) reconstruction through continuous exposure control.
Extensive experiments demonstrate that LUCID consistently outperforms state-of-the-art methods across diverse real-world nighttime scenarios.
\end{abstract}



\begin{teaserfigure}
  \includegraphics[width=\textwidth]{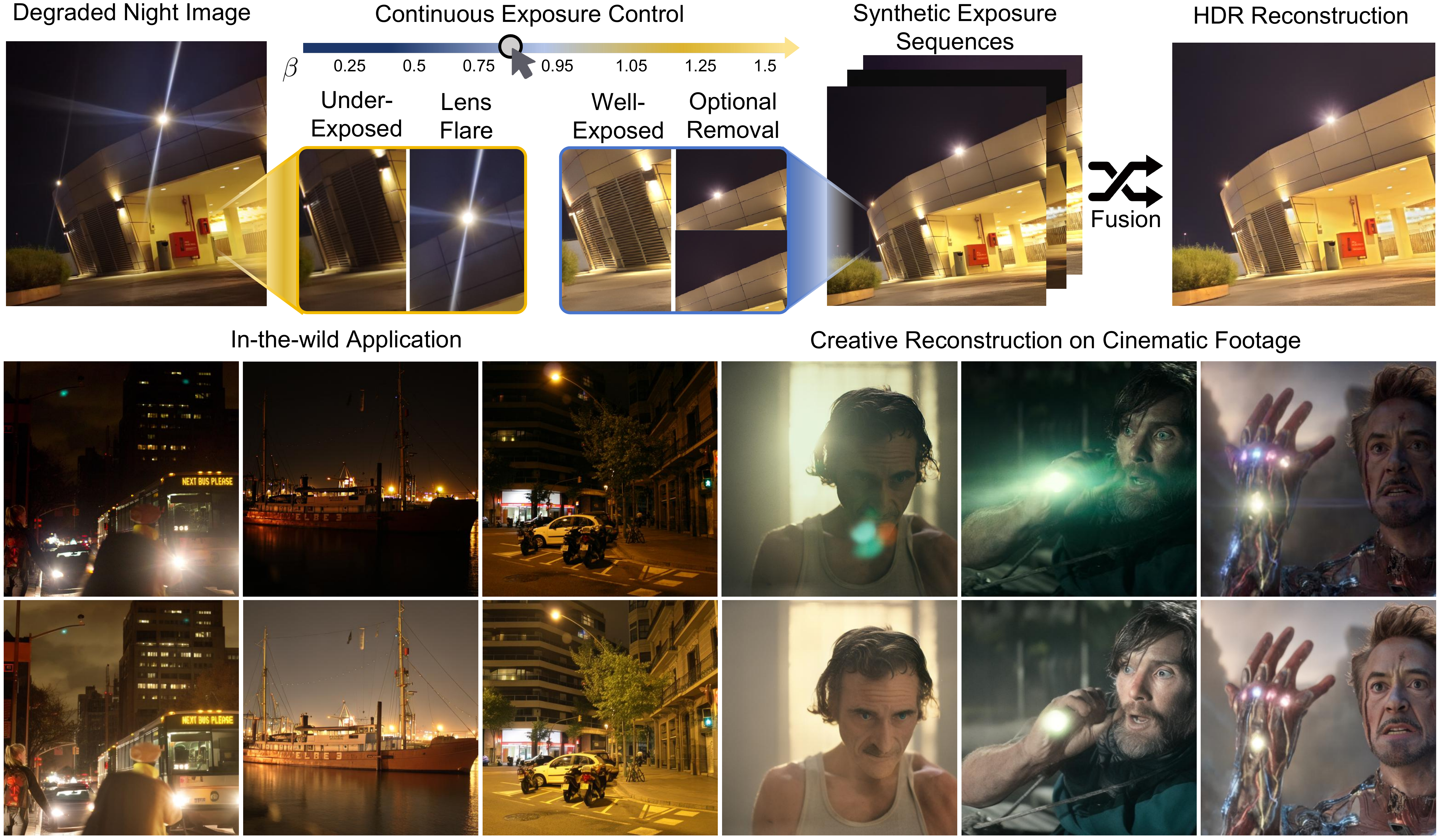}
  \caption{Unveiling the night with LUCID. We present a unified diffusion framework that jointly addresses severe underexposure and intense lens flares. By enabling continuous exposure modulation from a single view, LUCID facilitates the synthesis of pseudo-exposure sequences for HDR reconstruction (top). Bottom examples demonstrate robust generalization across complex in-the-wild illumination as well as cinematic footage, with film stills sourced from ShotDeck and for academic study of cinematic flare removal and editing. Project page: \textit{\url{https://xiaoyunyuan.net/index.html?project=lucid}}}
  \Description{A teaser figure showing LUCID restoring nighttime photographs with severe flare and underexposure, plus controlled exposure modulation for HDR-style reconstruction.}
  \label{fig:teaser}
\end{teaserfigure}

\maketitle

\section{Introduction}

Photography is the art of painting with light. Yet, at night, this canvas becomes notoriously difficult to control. While the human eye effortlessly adapts to the dance between deep shadows and piercing artificial lights, physical sensors struggle profoundly. Limited dynamic range swallows details in the dark, while bright sources spill into overwhelming flare, veiling the scene's true geometry. Admittedly, not all flares are unwanted; in nighttime cinematography, they are often deliberate signatures of mood.

In J.J. Abrams' \textit{Star Trek} (2009), for instance, horizontal anamorphic flares are embraced to evoke kinetic energy and a futuristic atmosphere. Conversely, maintaining optical purity requires immense physical effort. In the production of \textit{The Batman} (2022), the cinematography team employed massive physical barriers and custom-built lenses specifically to shield the sensor from stray city lights, fighting to preserve the deep, immersive blacks essential to its noir aesthetic. Fortunately, digital tools have fundamentally reshaped this landscape. Sculpting light in post-production is now routine, granting artists unprecedented creative freedom. Yet, a crucial asymmetry persists: while adding stylized flare to a clean image is straightforward, excavating a pristine signal from a glare-compromised capture is a formidable challenge, as the artifacts overwrite the very information needed for restoration. This motivates our goal: to computationally recover a clean, high-dynamic-range baseline that decouples creative intent from optical accidents, providing a reliable foundation for artistic night photography.

The challenges inherent to this domain, however, are fundamentally structural and intricate. Nighttime degradations are not mere suboptimal exposures, but a profound eclipse of information: intense flare bleeds across the sensor, erasing underlying geometry, while photon-starved regions dissolve into noise and quantization. Critically, restoration becomes a battle between opposing forces. Taming the flare risks extinguishing genuine highlights, while pulling details from the dark inevitably amplifies artifacts and residual ghosting. Thus, nighttime imaging is less a single restoration task and more a delicate balancing act among competing physical constraints.

\begin{figure}[t]
  \centering
  \includegraphics[width=\linewidth]{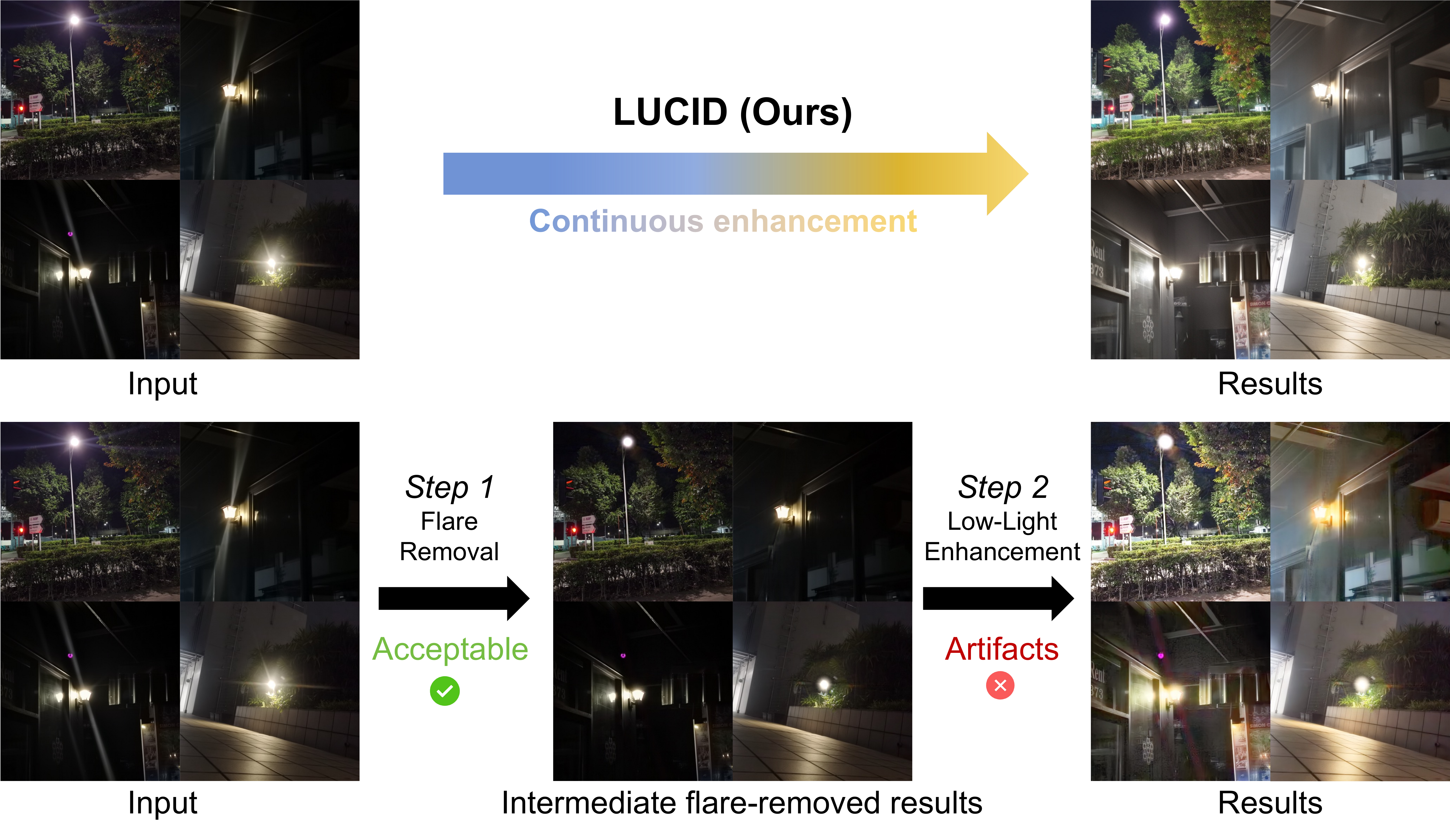}
  \caption{Comparison with a disjoint baseline. The bottom row illustrates the error accumulation inherent in sequential processing: while the intermediate flare-removed results (Step 1) appear acceptable, latent flare residuals are masked by darkness and subsequently amplified into severe artifacts during enhancement (Step 2). In contrast, our unified LUCID (top) jointly addresses both degradations, yielding clean and coherent reconstruction.}
  \Description{A comparison figure contrasting a sequential deflare-then-enhance pipeline with the unified LUCID model, highlighting artifact amplification in the disjoint baseline.}
  \label{Fig:artifacts}
\end{figure}

Most existing methods, however, treat low-light enhancement (LLIE) and flare mitigation as independent problems~\cite{cai2023retinexformer,Feijoo_2025_CVPRdarkir,dai2022flare7k,jiang2024mfdnet}, ignoring their physically coupled nature. This decoupling is perilous: simply cascading an LLIE model with a flare-removal method often yields unstable behaviors and severe artifacts (Fig.~\ref{Fig:artifacts}), as the errors from one stage are amplified by the other. Compounding this difficulty is the scarcity of dataset. Acquiring paired nighttime samples in the wild is hindered by the prohibitive difficulty of capturing clean references and the intrinsic ambiguity of ground-truth illumination. Consequently, current datasets largely rely on staged exposure bracketing~\cite{li2021benchmarkinglolv2,wei2018deeplolv1}, which restricts scene diversity and fails to capture the chaotic illumination patterns of real environments. Finally, most frameworks operate as rigid black boxes, lacking mechanisms for user agency. This prevents users from tailoring brightness or visibility to their needs.

To navigate this complex landscape, we draw inspiration from two pivotal advancements. First, the emergence of synthetic flare datasets, such as Flare7K~\cite{dai2022flare7k}, has offered a modular way to represent optical artifacts. By decomposing flare into additive components, it enables the flexible synthesis of training data that covers diverse scattering patterns. Simultaneously, diffusion models have reshaped image restoration~\cite{lin2024diffbir,wu2024seesr,wu2024osediff,2024s3diff} by providing powerful generative priors. Their innate ability to generate natural details and reconstruct missing structures makes them uniquely suited for creating a clean image from heavily degraded inputs. Nevertheless, how to integrate these generic priors to handle the specific, adversarial constraints of nighttime photography remains an open question.

Building on these insights, we introduce LUCID: a unified framework for continuous flare mitigation and exposure adjustment in nighttime photography (Fig.~\ref{Fig:pipeline}). Instead of treating restoration as a static regression, LUCID decouples the problem into two distinct yet cooperative stages. First, a Flare Disentanglement Module isolates the ``curtain'' of light artifacts, extracting a structural flare map that serves as a precise guide for the subsequent process. Second, a Diffusion-Driven Restoration Module incorporates reference mixing layers to reorganize this disentangled information, reconstructing a clean, well-exposed image. Crucially, to empower the creator, we introduce a novel four-mode training strategy. By supervising the model with stratified pairings of exposure and light source intensity, we enable continuous control during inference. Through classifier-free guidance (CFG), users can smoothly control the restoration process, adjusting the output from a flare-dominated input to a clean, well-balanced image within a single unified model.

Finally, LUCID redefines the workflow of nighttime photography, supporting fine-grained control ranging from fully removing flare to preserving the natural structure of light sources. Extensive experiments demonstrate that LUCID not only produces visually superior results compared to state-of-the-art (SOTA) baselines but also exhibits robust generalization to diverse real-world scenes. Beyond standard restoration, LUCID naturally extends to High-Dynamic-Range (HDR) reconstruction, recovering faithful luminance transitions from single exposures. By bridging the gap between physical limitations and creative intent, LUCID offers a versatile instrument for both automated enhancement and artistic expression.

\section{Related Works}

\subsection{Low-Light and Nighttime Image Enhancement}

Deep learning has significantly advanced low-light image enhancement (LLIE), evolving from early CNN-based decomposition~\cite{wei2018deeplolv1,zhang2019kindling} to more advanced restoration frameworks. Zero-DCE~\cite{Zero-DCE} removes the need for paired data via a zero-reference curve estimation strategy. Malik and Soundararajan~\shortcite{malik2023semi} leverage quality-assisted pseudo-labels to reduce reliance on paired data. Subsequent methods focus on global context and degradation coupling. Retinexformer~\cite{cai2023retinexformer} models long-range illumination dependencies, DarkIR~\cite{Feijoo_2025_CVPRdarkir} addresses coupled low-light and blur, and Reti-Diff~\cite{he2025retidiff} incorporates diffusion process to enhance perceptual quality. LUXFormer~\cite{zhou2025luxformer} enhances illumination modeling via spatial-frequency priors.
Enhancing dark regions often amplifies glow and saturates light sources~\cite{sharma2021nighttime}. Recent works attempt to balance this trade-off or explore unified modeling for these coupled degradations~\cite{jin2022unsupervised,ren2025sfrbe,wu2023generation,jin2023enhancing,bernabel2024ndels}. While existing methods improve visibility, their performance in real-world nighttime scenes is limited by complex illumination conditions and a lack of controllability, particularly over light sources.

\subsection{Nighttime Flare Mitigation}

Nighttime flare mitigation presents a formidable challenge due to the high dynamic range of artificial lights and the intricate scattering patterns they produce. While hardware solutions like anti-reflective coatings~\cite{macleod2010thin} and fluid-filled lenses~\cite{boynton2003liquid} offer physical suppression, they struggle to fully eliminate artifacts under extreme contrast. Data-driven approaches emerge to address such problem. Wu \textit{et al.}~\shortcite{wu2021train} pioneered the use of optical Point Spread Functions (PSF) for semi-synthetic training, a strategy refined by Flare7K~\cite{dai2022flare7k} using real-world statistical priors. To enhance physical realism, Zhou \textit{et al.}~\shortcite{zhou2023improving} further model ISP and automatic exposure physics to better recover saturated sources. MfdNet~\cite{jiang2024mfdnet} advances the field by leveraging frequency domain analysis. Additionally, the physical constraints of compact mobile optics introduce highly specific sources of flare. This has spurred targeted research addressing artifacts originating from under-display camera diffraction~\cite{feng2021removing,song2023under,ahn2025udc} and smartphone-specific reflective surfaces~\cite{dai2023nighttime}. However, intense flares often cause information loss by completely occluding underlying scene structures. This necessitates the use of generative models capable of inferring missing content.

\subsection{Diffusion Models}

Recent advances have repurposed the rich generative capabilities of pre-trained diffusion models for image restoration. Frameworks like DiffBir~\cite{lin2024diffbir} and SeeSR~\cite{wu2024seesr} integrate structural and semantic guidance to steer the denoising trajectory, while SUPIR~\cite{yu2024supir} introduces partial controllability via Classifier-Free Guidance (CFG). To accelerate inference, methods such as Difix3D+~\cite{wu2025difix3d+} and DMDiff~\cite{zhang2025dmdiff} leverage one-step diffusion backbones, enabling correction of complex degradations ranging from metalens aberrations to 3D rendering errors. Unified frameworks that handle multiple degradation types within a single model have also emerged, leveraging language-conditioned~\cite{luo2024daclip,jiang2024autodir} and low-level cue-guided~\cite{mandal2025unicorn} diffusion priors. In the domain of HDR imaging, LEDiff~\cite{li2025lediff} and GaSLight~\cite{bolduc2025gaslight} perform latent space manipulation to estimate spatially-varying illuminance, plausibly reconstructing clipped highlights and shadows. Specifically for lens flare, Difflare~\cite{zhou2024difflare} exploits diffusion priors to hallucinate scene content occluded by saturated scattering artifacts.

However, current approaches overlook the inherently subjective nature of nighttime enhancement. The optimal balance between clarity and atmosphere is dictated by artistic intent rather than a fixed target, yet existing methods are constrained to deterministic mappings that cannot accommodate such aesthetic diversity. This motivates the need for continuous, user-steerable control.

\begin{figure*}[t]
  \centering
  \includegraphics[width=\textwidth]{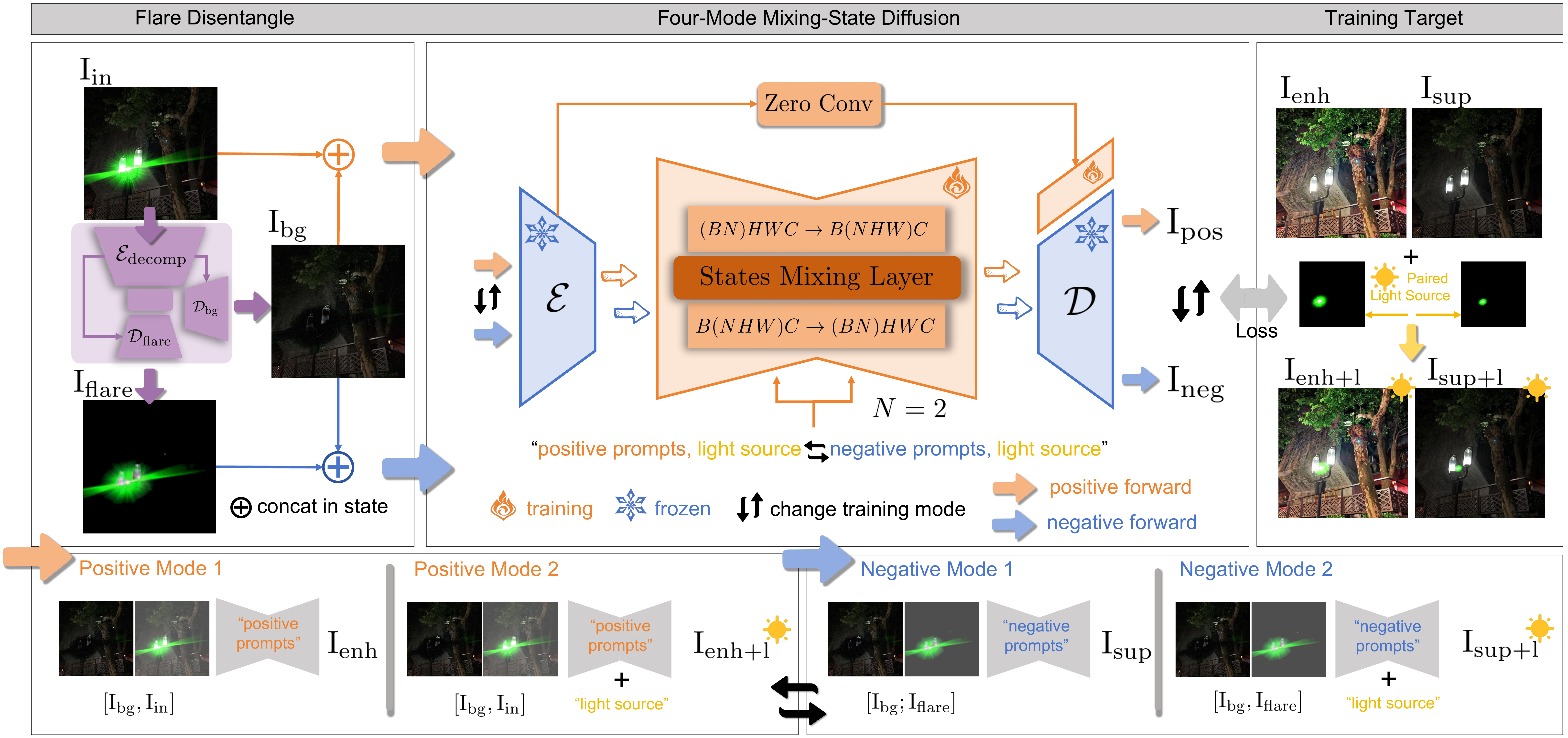}
  \caption{Training framework of LUCID. The pipeline begins with the Flare Disentangle stage (left), where the degraded input $\mathbf{I}_{in}$ is decomposed into a background estimate $\mathbf{I}_\text{bg}$ and a flare component $\mathbf{I}_\text{flare}$. These components are selectively concatenated along the state dimension to form multi-state inputs for the Mixing-state Diffusion (center). Inside the network, cross-state interactions guide the diffusion process to achieve fine-grained restoration. The bottom panel illustrates our four distinct training regimes: we alternate between positive modes (using $\mathbf{I}_\text{in}$ as reference for enhancement) and negative modes (using $\mathbf{I}_\text{flare}$ as reference for suppression), with optional ``light source'' prompts to enforce semantic controllability over high-intensity regions. A summary of notations is provided in the supplementary material to ensure clearer and more intuitive presentation of the framework.}
  \Description{A pipeline diagram of LUCID showing flare disentanglement, mixing-state diffusion, and four training modes for enhancement, suppression, and light-source control.}
  \label{Fig:pipeline}
\end{figure*}

\section{Methodology}

We view nighttime photography not merely as a pixel-level restoration problem, but as a re-lighting process that seeks to recover visual information jointly obscured by darkness and optical flare. From a physical perspective, nighttime degradation arises from the coexistence of insufficient photon exposure and strong stray light, resulting in both signal collapse and structured artifacts.

Following the Retinex formulation~\cite{land1971lightness}, which is widely adopted in nighttime imaging~\cite{yi2023diffretinex,cai2023retinexformer}, image formation is commonly modeled as \(I = R \cdot L\), where \(R\) and \(L\) denote the intrinsic scene reflectance and ambient illumination, respectively. In the presence of intense light sources, this model is naturally extended to \(I = R \cdot L + F\), where \(F\) represents additive stray light caused by flare and ghosting.

This formulation reveals two mathematically distinct restoration objectives: recovering multiplicative illumination and suppressing additive flare. However, in complex real nighttime scenes, these degradations are physically entangled. Naively increasing exposure amplifies flare artifacts, while aggressively suppressing flare often destroys legitimate scene structure. This inherent tension motivates a framework that decomposes these factors explicitly, yet resolves them cooperatively (Fig.~\ref{Fig:pipeline}).

\subsection{Flare Disentanglement}

We begin by explicitly disentangling additive optical artifacts from the underlying scene content. Observing that flare exhibits strong spatial coherence, such as smooth halos and structured ghosting patterns, we employ a lightweight U-Net with a shared encoder and two parallel decoders to decompose the input nighttime image \(\mathbf{I}_{\text{in}}\) into a flare component \(\mathbf{I}_{\text{flare}}\) and a background component \(\mathbf{I}_{\text{bg}}\):
\begin{equation}
\begin{aligned}
\mathbf{I}_{\text{flare}} &= \mathcal{D}_{\text{flare}}\!\left(\mathcal{E}_{\text{decomp}}(\mathbf{I}_{\text{in}})\right), \\
\mathbf{I}_{\text{bg}} &= \mathcal{D}_{\text{bg}}\!\left(\mathcal{E}_{\text{decomp}}(\mathbf{I}_{\text{in}})\right).
\end{aligned}
\end{equation}
Here, \(\mathcal{E}_{\text{decomp}}\) denotes the shared encoder, while \(\mathcal{D}_{\text{flare}}\) and \(\mathcal{D}_{\text{bg}}\) are the flare and background decoders, respectively. Rather than explicitly separating reflectance \(R\) and illumination \(L\), we use the term ``background'' to represent their product \(R \cdot L\), which provides a structurally faithful yet flare-free representation. This decomposition serves two complementary purposes: isolating additive flare artifacts for targeted suppression and producing a reliable structural prior to guide the subsequent restoration.

To encourage meaningful disentanglement, the two decoders share weights for the first \(m\) layers and diverge thereafter. We impose an orthogonality constraint on the feature maps at the \(k\)th layer (first divergent layer) to enforce mutual exclusivity:
\begin{equation}
\mathcal{L}_{\text{ortho}} = \left\| \mathbf{F}^{k}_{\text{bg}} \odot \mathbf{F}^{k}_{\text{flare}} \right\|_2^2,
\end{equation}
where \(\mathbf{F}^{k}_{\text{bg}}\) and \(\mathbf{F}^{k}_{\text{flare}}\) denote the feature maps at the \(k\)th layer of the respective decoders. In addition, we impose a reconstruction constraint to ensure self-consistency:
\begin{equation}
\mathcal{L}_{\text{recon}} =
\left\| (\mathbf{I}_{\text{flare}} + \mathbf{I}_{\text{bg}}) - \mathbf{I}_{\text{in}} \right\|_2^2.
\end{equation}
Combined with component-level supervision, the flare disentanglement network is trained using reconstruction, orthogonality, and standard $\ell_2$ losses on the flare and background components:
\begin{equation}
\mathcal{L} = \mathcal{L}_{\text{recon}} + \mathcal{L}_{\text{ortho}} + \mathcal{L}_{\text{comp}},
\end{equation}
where $\mathcal{L}_{\text{comp}}$ denotes the $\ell_2$ supervision on the predicted flare and background against their ground-truth counterparts.

\subsection{Four-Mode Mixing-State Diffusion}

Although flare mitigation and illumination recovery are decomposed at the representation level, they must ultimately be resolved jointly to ensure holistic visual consistency. To this end, we introduce a diffusion-based restoration module that explicitly models their interaction. Drawing architectural inspiration from the paradigm of multi-view diffusion~\cite{wu2025difix3d+,shi2023MVDream}, we cast the restoration as a reconstruction process driven by dual priors: \(\{\mathbf{I}_{\text{main}}, \mathbf{I}_{\text{ref}}\} \rightarrow \mathbf{I}_{\text{out}}\), where the reference image provides auxiliary states for structure or flare characteristics. Both inputs are encoded into the latent space via a shared VAE encoder \(\mathcal{E}\):
\begin{equation}
\mathbf{z}_{\text{main}} = \mathcal{E}(\mathbf{I}_{\text{main}}), \quad
\mathbf{z}_{\text{ref}} = \mathcal{E}(\mathbf{I}_{\text{ref}}).
\end{equation}

Cross-state interaction is realized through mixing-state self-attention layers embedded within the denoising U-Net. Prior to each attention operation, the latent representations are concatenated along a state dimension, \(\mathbf{z} \in \mathbb{R}^{B \times 2 \times C \times H \times W}\), then rearranged to collapse the state dimension into the spatial domain, enabling attention across both inputs:
\begin{equation}
\begin{aligned}
\mathbf{z}' &= \text{Rearrange}(\mathbf{z}, B \times (2HW) \times C), \\
\mathbf{z}' &= \text{Self-Attention}(\mathbf{z}', \mathbf{z}'), \\
\mathbf{z}' &= \text{Rearrange}(\mathbf{z}', B \times 2 \times C \times H \times W).
\end{aligned}
\end{equation}
This mechanism compels the network to explicitly attend to cues provided by the reference state, enabling robust extraction of structure- and flare-related priors under a unified denoising process.

To further constrain illumination fidelity, we reuse the encoder of the flare disentanglement network as a semantic feature extractor and define an intrinsic feature loss:
\begin{equation}
\mathcal{L}_{\text{intri}} =
\sum_{\ell} w_{\ell}
\left\| \mathbf{f}^{\ell}_{\text{out}} - \mathbf{f}^{\ell}_{\text{tgt}} \right\|_2^2,
\end{equation}
where $\mathbf{f}^{\ell}_{\text{out}}$ and $\mathbf{f}^{\ell}_{\text{tgt}}$ denote the $\ell$-th layer features extracted from the network output $\mathbf{I}_{out}$ and the corresponding training target $\mathbf{I}_{tgt}$, respectively. The final diffusion loss combines pixel-level and perceptual objectives:
\begin{equation}
\mathcal{L}_{\text{diff}} =
\mathcal{L}_{\text{intri}} + \mathcal{L}_2 + \mathcal{L}_{\text{LPIPS}}.
\end{equation}

\begin{figure}[t]
  \centering
  \begin{subfigure}{0.32\linewidth}
    \includegraphics[width=\linewidth]{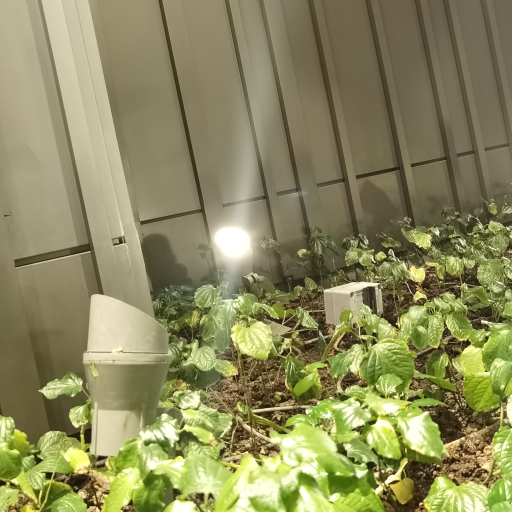}
    \caption{Input}
  \end{subfigure}
  \hfill
  \begin{subfigure}{0.32\linewidth}
    \includegraphics[width=\linewidth]{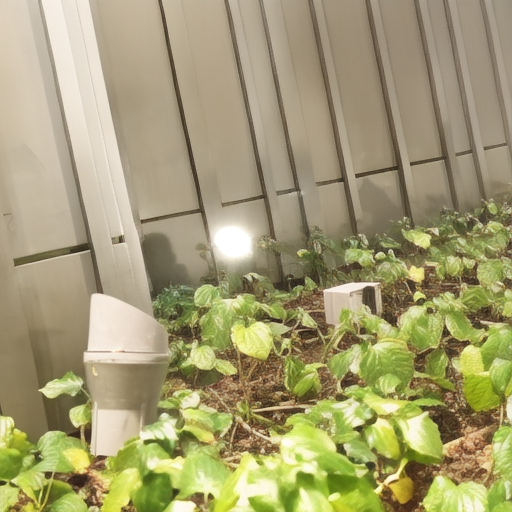}
    \caption{``Light Source''}
  \end{subfigure}
  \hfill
  \begin{subfigure}{0.32\linewidth}
    \includegraphics[width=\linewidth]{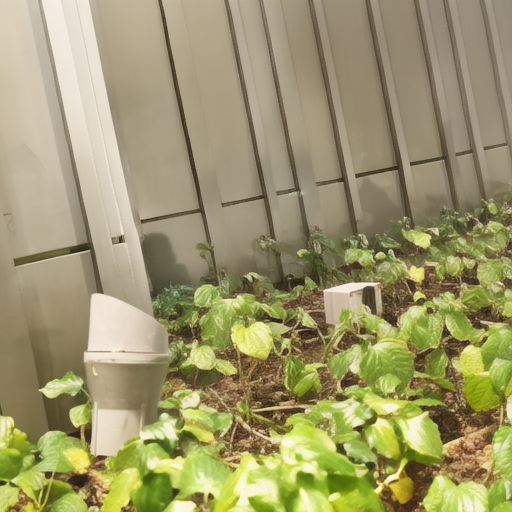}
    \caption{w/o ``Light Source''}
  \end{subfigure}
  \caption{Prompt-driven light source preservation. By supervising the network with distinct GT targets conditioned on specific textual prompts, LUCID learns to selectively retain or fully remove the light source.}
  \Description{A three-panel comparison showing the input image, restoration with a light-source prompt, and restoration without the prompt to demonstrate controllable source preservation.}
  \label{fig:prompt}
\end{figure}

\subsection{Classifier-Free Guidance-Based Control}

Classifier-Free Guidance (CFG)~\cite{Ho2022ClassifierFreeDG} provides a principled mechanism for conditional control in diffusion models. While prior restoration methods~\cite{2024s3diff,zhang2025dmdiff} primarily use CFG to trade off fidelity and realism, we extend this paradigm to enable continuous control over exposure and light source appearance. As illustrated in Fig.~\ref{Fig:pipeline}, we strategically design four training modes defined by different combinations of inputs, references, and targets, to support continuous control along two orthogonal dimensions: exposure and light-source presence.

\textit{Continuous Exposure Control.} To model exposure as a controllable dimension, we define two complementary training modes. In the positive mode, the main input $\mathbf{I}_{\text{main}}$ is the entangled background \(\mathbf{I}_{\text{bg}}\), the reference is the original nighttime image \(\mathbf{I}_{\text{in}}\), and the target is a flare-free, well-exposed image \(\mathbf{I}_{\text{enh}}\). This configuration encourages the model to exploit reference-derived structural cues for both illumination recovery and flare suppression. In contrast, the negative mode replaces the reference with the flare \(\mathbf{I}_{\text{flare}}\) and sets the target to an under-exposed image \(\mathbf{I}_{\text{sup}}\), guiding the model to associate the reference state with flare characteristics alone. These two modes jointly define the endpoints of the exposure control spectrum.

\textit{Continuous Light-Source Control.} To further empower selective manipulation of scene illuminants, we augment each exposure mode with specialized sub-configurations designed to either preserve or suppress explicit light-source appearances. We synthesize pseudo-illuminant maps, $\mathbf{I}_{\text{enh+l}}$ and $\mathbf{I}_{\text{sup+l}}$, adapting the synthesis protocols of Flare7K. Crucially, these components are engineered to simulate the pristine, intrinsic radiance of light sources devoid of optical scattering defects. For the supervision of these light-preserving sub-modes, we designate the composite states as the GT targets: integrating $\mathbf{I}_{\text{enh+l}}$ into $\mathbf{I}_{\text{enh}}$ for the positive regime, and $\mathbf{I}_{\text{sup+l}}$ into $\mathbf{I}_{\text{sup}}$ for the negative counterpart, respectively. To explicitly trigger this modal control, we append the textual descriptor ``light source'' to the base prompts. This instructs the model to semantically modulate the restoration behavior, ensuring the faithful retention of illuminants aligned with the constructed targets.

At inference time, CFG enables continuous traversal between negative and positive solutions by modulating a scalar \(\beta\):
\begin{equation}
\hat{\mathbf{z}} = \mathbf{z}_{\text{neg}} + \beta (\mathbf{z}_{\text{pos}} - \mathbf{z}_{\text{neg}}).
\end{equation}
After latent interpolation, a VAE decoder augmented with encoder skip connections reconstructs the final output. By adjusting \(\beta\) and toggling the ``light source'' prompt, users can seamlessly transition from aggressive flare suppression to selective preservation of light sources and their associated artifacts.

\subsection{Nighttime Single-Image HDR Reconstruction}

The proposed framework naturally extends to single-image High Dynamic Range (HDR) reconstruction. By leveraging the continuous exposure control enabled by CFG, a sequence of exposure-consistent outputs can be synthesized from a single nighttime input. These outputs are fused to produce the final HDR result. We employ Laplacian pyramid blending with quality-aware weighting to aggregate the synthesized exposures into an HDR representation with balanced luminance distribution and light-source appearance. Detailed algorithmic procedures are provided in the Supplementary Material.

\begin{figure}[t]
  \centering
  \includegraphics[width=\linewidth]{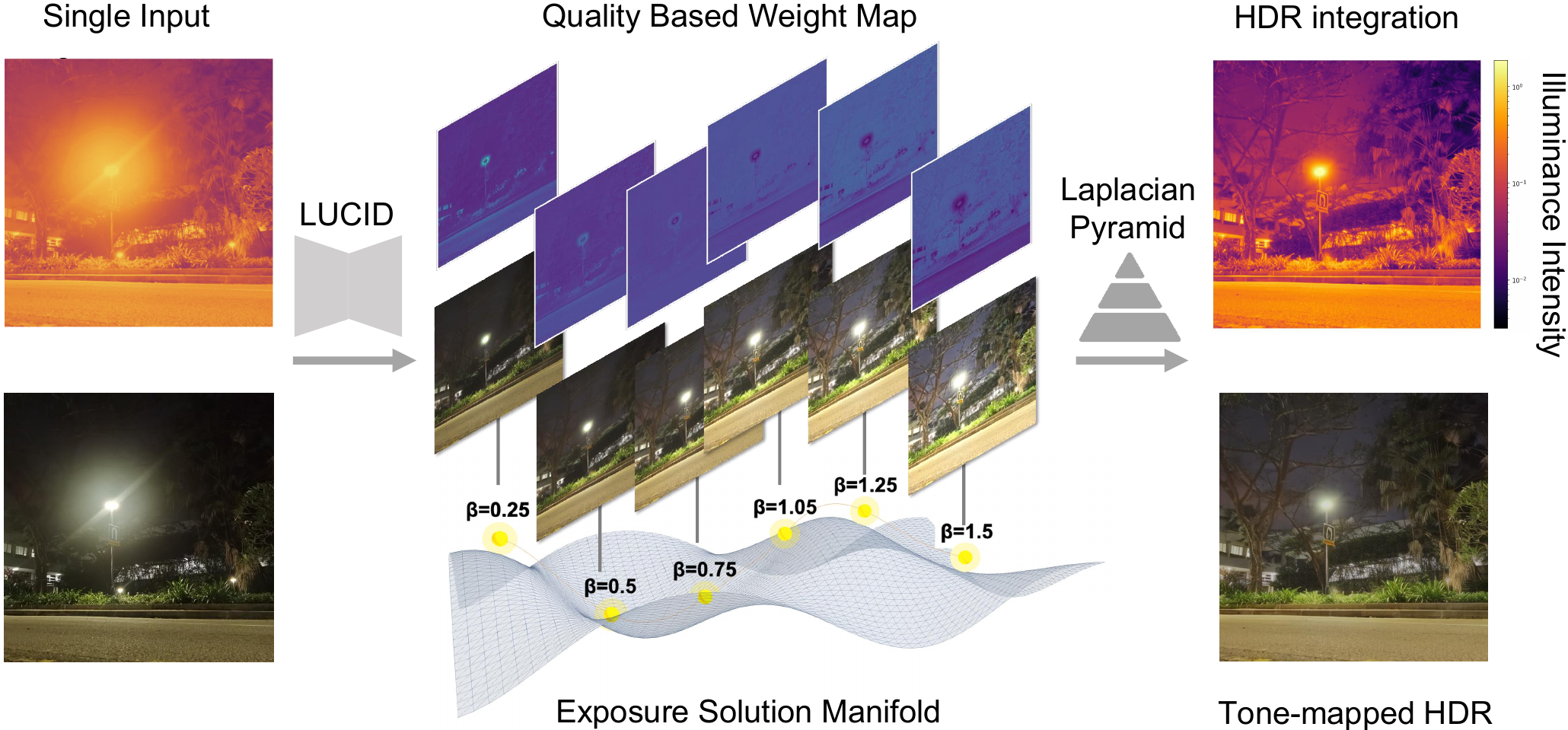}
  \caption{Single-image HDR reconstruction.}
  \Description{A diagram illustrating LUCID synthesizing multiple virtual exposures from one nighttime image and merging them into an HDR reconstruction.}
  \label{Fig:manifold}
\end{figure}

\section{Experiments and Results}

\subsection{Implementation}

We adapt SD-Turbo~\cite{sauer2023adversarial} as the generative backbone of our diffusion-based restoration module. Training data are curated from several real-world low-light datasets, including RELLISUR~\cite{aakerberg2021rellisur}, LSRW~\cite{hai2023r2rnet}, SICE~\cite{Cai2018sice}, and SID~\cite{chen2018learningsid}. To synthesize degraded nighttime inputs \(\mathbf{I}_{\text{in}}\), we follow the physically grounded flare generation pipeline of Flare7K~\cite{dai2022flare7k}, superimposing optical flare artifacts onto under-exposed images \(\mathbf{I}_{\text{sup}}\). During training, negative conditioning in CFG is applied with a probability of \(0.2\). Independently, the presence of the light-source semantic prompt is sampled with a probability of \(0.5\). Training is conducted on a single NVIDIA A800 GPU at a spatial resolution of \(512 \times 512\) with a batch size of \(4\).

\subsection{Evaluation Protocol}

\subsubsection{Evaluation Benchmark.} 
To evaluate LUCID under realistic nighttime photography conditions, we adapt three task-oriented complementary benchmarks.

\paragraph{General nighttime enhancement:} 
We build a diverse benchmark based on the Exclusively Dark (ExDark) dataset~\cite{Exdark}, which contains 7,363 real-world nighttime images ranging from extreme darkness to twilight. Unlike controlled low-light datasets, ExDark exhibits strong illumination imbalance and complex light-source interference that better reflect real nighttime photography. To exclude samples compromised by severe degradations (e.g., heavy blur or statistical distortions), we applied a quality screening process based on Laplacian variance and BRISQUE~\cite{mittal2012nobrisque} scores. Consequently, we select a subset of 1,271 images that feature pronounced dynamic range, visible light sources, and minimal ambient illumination for reliable perceptual evaluation.

\paragraph{Flare mitigation:} 
We use the Flare7K dataset~\cite{dai2022flare7k}, which provides real-world nighttime images containing prominent lens flare and ghosting artifacts.

\paragraph{Single-image HDR:} We include the SiHDR~\cite{hanji_mantiuk_eilertsen_hajisharif_unger_2022sihdr} dataset to assess the effectiveness of LUCID in recovering extended dynamic range from a single nighttime exposure.

\subsubsection{Comparison Methods.} 
We compare LUCID against recent state-of-the-art (SOTA) low-light image enhancement (LLIE) methods. To comprehensively benchmark general enhancement performance, we select representative methods spanning diverse technical paradigms, including Zero-DCE~\cite{Zero-DCE}, RetinexFormer~\cite{cai2023retinexformer}, RetiDiff~\cite{he2025retidiff}, DarkIR~\cite{Feijoo_2025_CVPRdarkir}, and the unsupervised approach of Jin \textit{et al.}~\shortcite{jin2022unsupervised}. For all competing approaches, we use the official pre-trained models recommended by the authors for in-the-wild inference to ensure a fair comparison under consistent experimental settings.

To evaluate flare mitigation, we additionally include representative flare removal methods, namely Flare7K~\cite{dai2022flare7k}, MFDNet~\cite{jiang2024mfdnet}, and the method of Zhou \textit{et al.}~\shortcite{zhou2023improving}. For single-image HDR reconstruction, we compare against established HDR methods, including IntrinsicHDR~\cite{dilleIntrinsicHDR}, LEDiff~\cite{li2025lediff}, and GasLight~\cite{bolduc2025gaslight}.

Unlike controlled low-light benchmarks, which are constructed by varying camera exposure and therefore exhibit limited dynamic range, real nighttime scenes involve highly heterogeneous illumination with no unique ground-truth brightness. Accordingly, we primarily rely on no-reference image quality metrics~\cite{wang2022clipiqa,yang2022maniqa,ke2021musiq,zhang2023liqe,table8352823nima} for quantitative evaluation.

\newcommand{\sixcolgap}{\hspace{0.006\linewidth}}
\newcommand{\fivecolgap}{\hspace{0.006\linewidth}}

\begin{figure*}[p]
  \centering
  \makebox[\linewidth][c]{%
  \begin{subfigure}[b]{0.136\linewidth}
    \includegraphics[width=\linewidth]{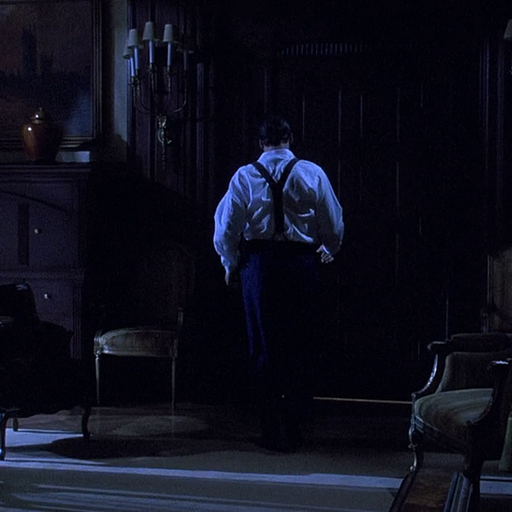}
  \end{subfigure}
  \sixcolgap
  \begin{subfigure}[b]{0.136\linewidth}
    \includegraphics[width=\linewidth]{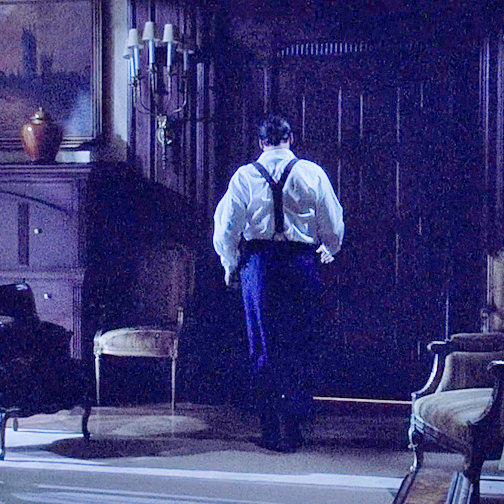}
  \end{subfigure}
  \sixcolgap
  \begin{subfigure}[b]{0.136\linewidth}
    \includegraphics[width=\linewidth]{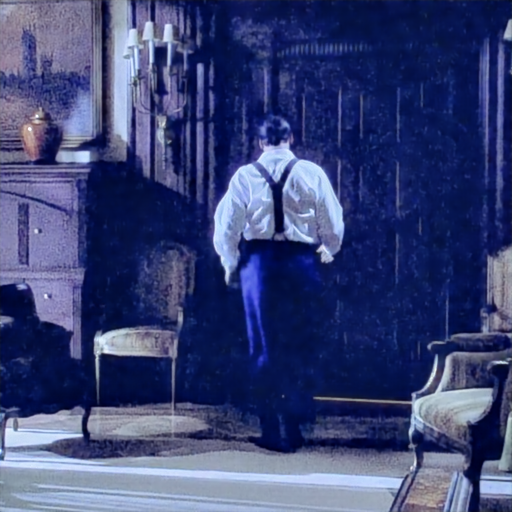}
  \end{subfigure}
  \sixcolgap
  \begin{subfigure}[b]{0.136\linewidth}
    \includegraphics[width=\linewidth]{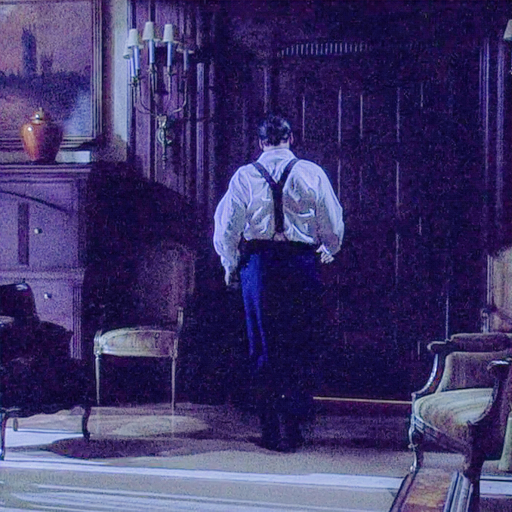}
  \end{subfigure}
  \sixcolgap
  \begin{subfigure}[b]{0.136\linewidth}
    \includegraphics[width=\linewidth]{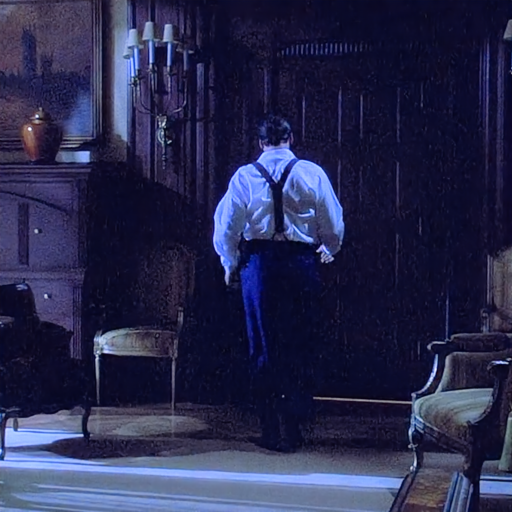}
  \end{subfigure}
  \sixcolgap
  \begin{subfigure}[b]{0.136\linewidth}
    \includegraphics[width=\linewidth]{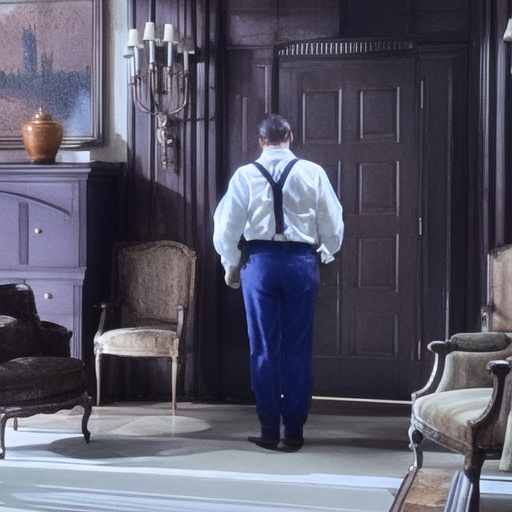}
  \end{subfigure}}

  \par\smallskip

  \makebox[\linewidth][c]{%
  \begin{subfigure}[b]{0.136\linewidth}
    \includegraphics[width=\linewidth]{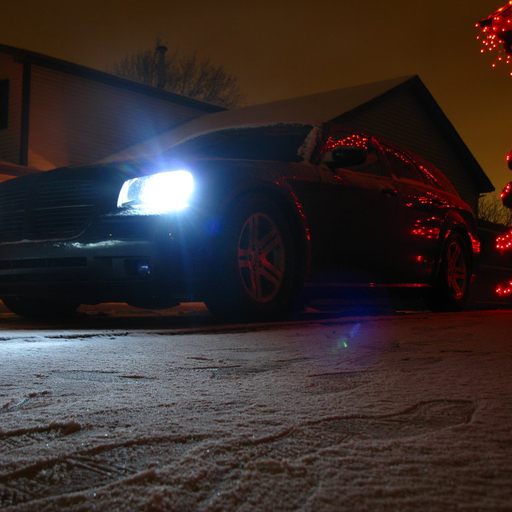}
  \end{subfigure}
  \sixcolgap
  \begin{subfigure}[b]{0.136\linewidth}
    \includegraphics[width=\linewidth]{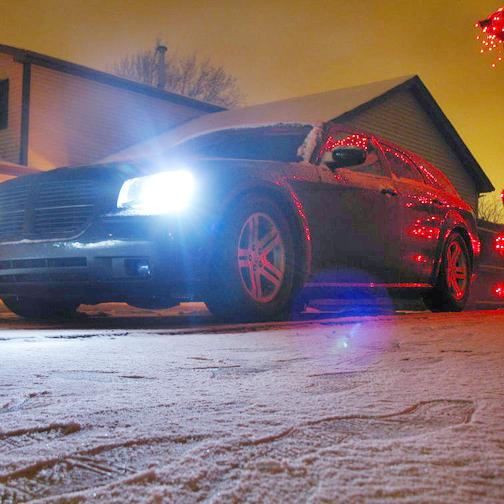}
  \end{subfigure}
  \sixcolgap
  \begin{subfigure}[b]{0.136\linewidth}
    \includegraphics[width=\linewidth]{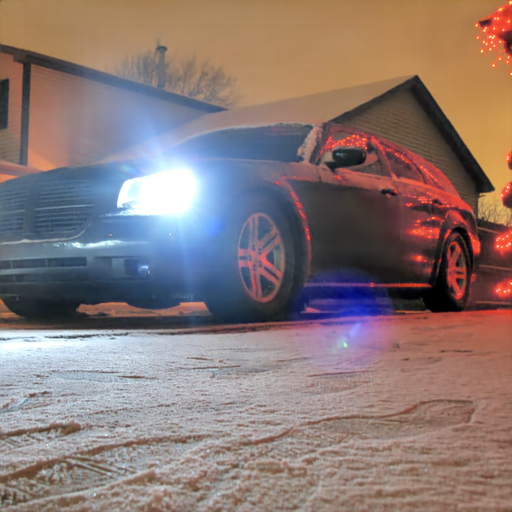}
  \end{subfigure}
  \sixcolgap
  \begin{subfigure}[b]{0.136\linewidth}
    \includegraphics[width=\linewidth]{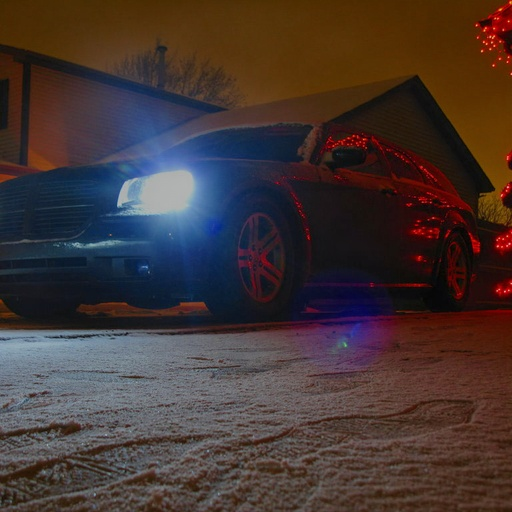}
  \end{subfigure}
  \sixcolgap
  \begin{subfigure}[b]{0.136\linewidth}
    \includegraphics[width=\linewidth]{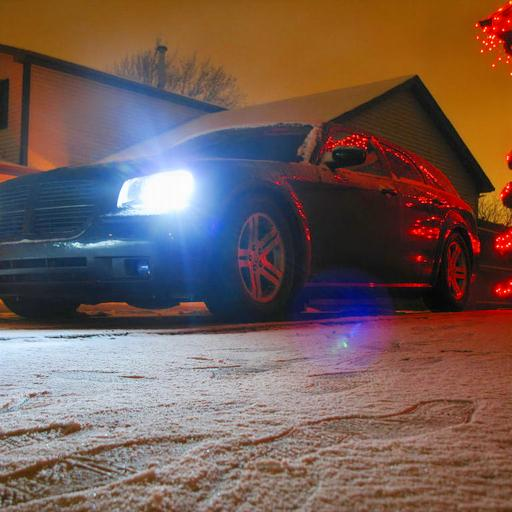}
  \end{subfigure}
  \sixcolgap
  \begin{subfigure}[b]{0.136\linewidth}
    \includegraphics[width=\linewidth]{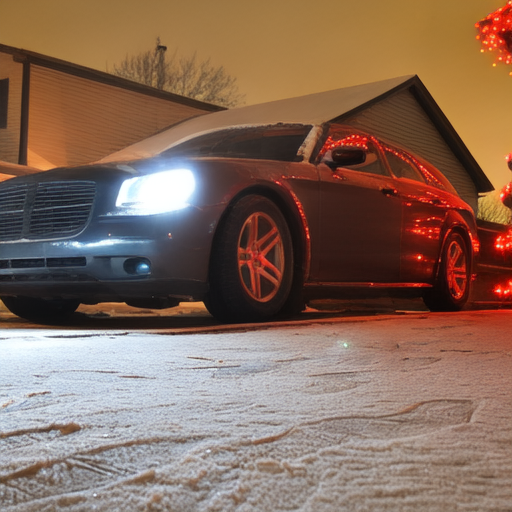}
  \end{subfigure}}

  \par\smallskip

  \makebox[\linewidth][c]{%
  \begin{subfigure}[b]{0.136\linewidth}
    \includegraphics[width=\linewidth]{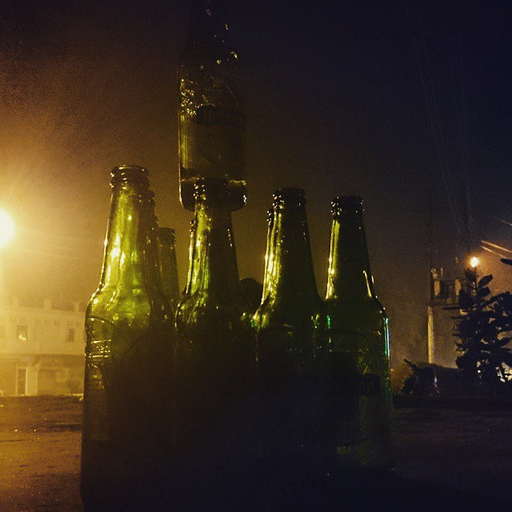}
  \end{subfigure}
  \sixcolgap
  \begin{subfigure}[b]{0.136\linewidth}
    \includegraphics[width=\linewidth]{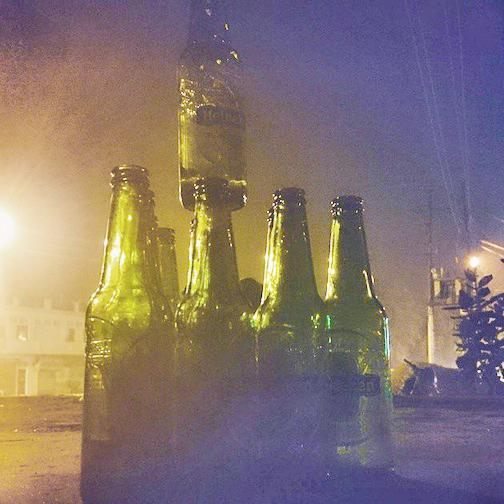}
  \end{subfigure}
  \sixcolgap
  \begin{subfigure}[b]{0.136\linewidth}
    \includegraphics[width=\linewidth]{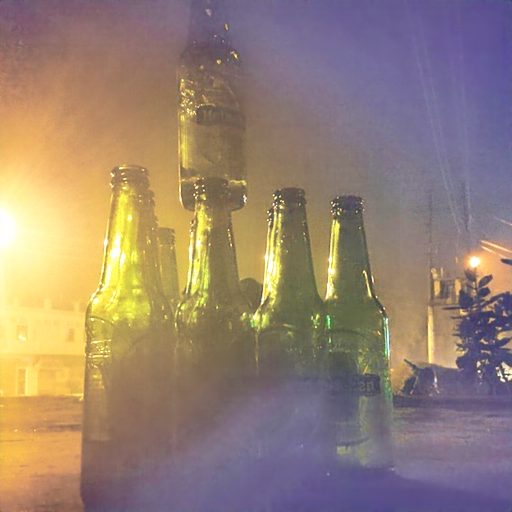}
  \end{subfigure}
  \sixcolgap
  \begin{subfigure}[b]{0.136\linewidth}
    \includegraphics[width=\linewidth]{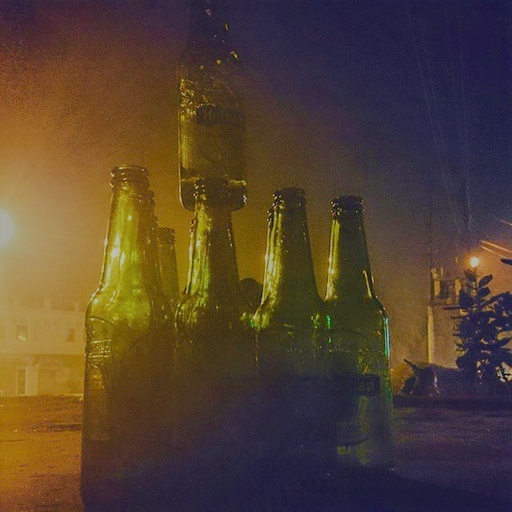}
  \end{subfigure}
  \sixcolgap
  \begin{subfigure}[b]{0.136\linewidth}
    \includegraphics[width=\linewidth]{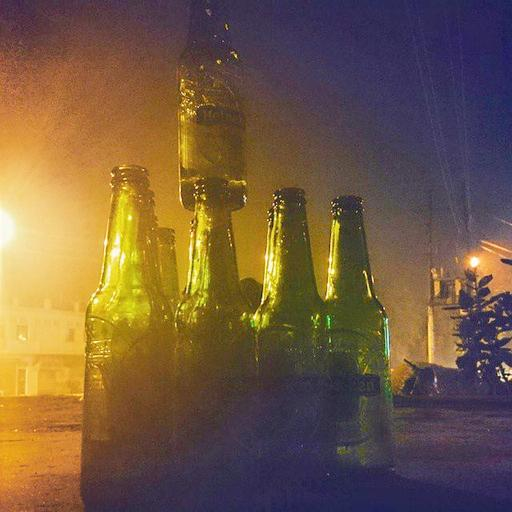}
  \end{subfigure}
  \sixcolgap
  \begin{subfigure}[b]{0.136\linewidth}
    \includegraphics[width=\linewidth]{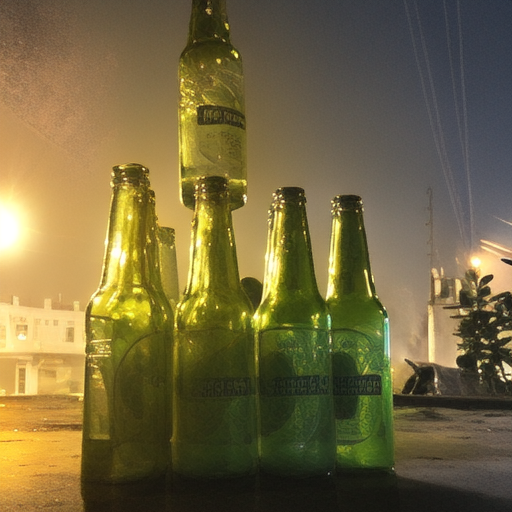}
  \end{subfigure}}

  \par\smallskip

  \makebox[\linewidth][c]{%
  \begin{subfigure}[b]{0.136\linewidth}
    \includegraphics[width=\linewidth]{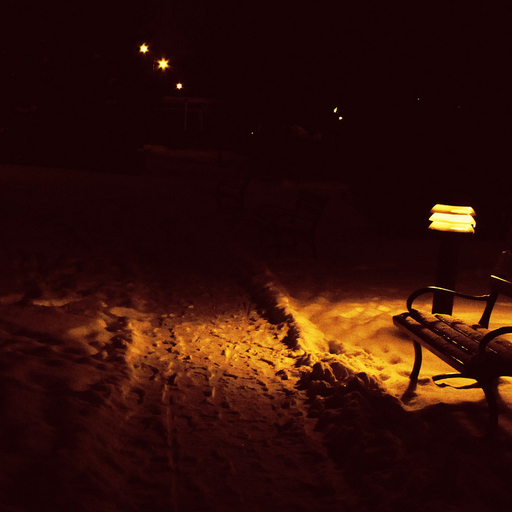}
  \end{subfigure}
  \sixcolgap
  \begin{subfigure}[b]{0.136\linewidth}
    \includegraphics[width=\linewidth]{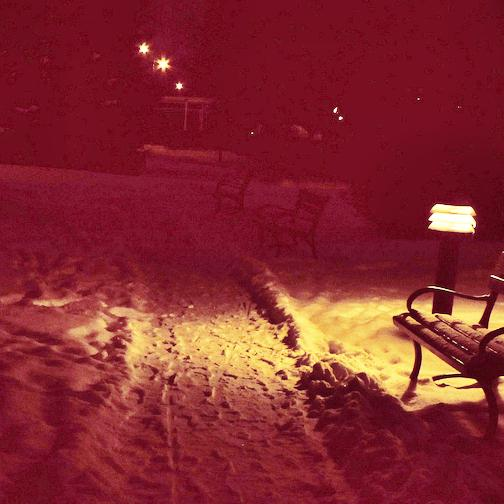}
  \end{subfigure}
  \sixcolgap
  \begin{subfigure}[b]{0.136\linewidth}
    \includegraphics[width=\linewidth]{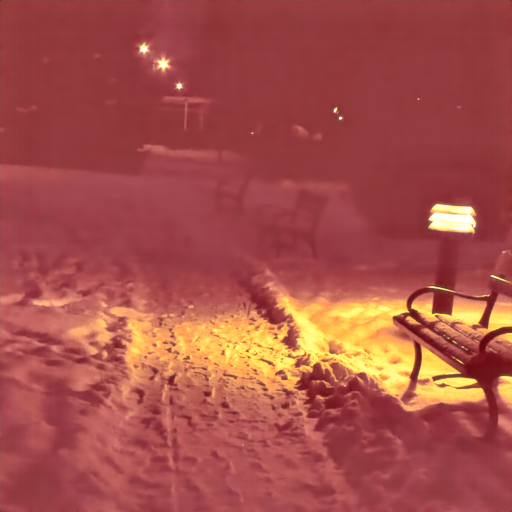}
  \end{subfigure}
  \sixcolgap
  \begin{subfigure}[b]{0.136\linewidth}
    \includegraphics[width=\linewidth]{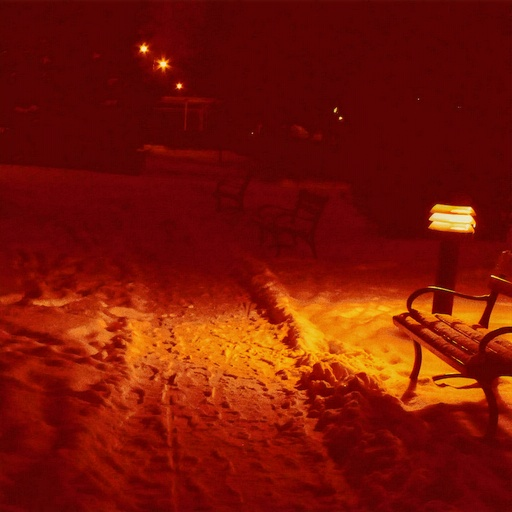}
  \end{subfigure}
  \sixcolgap
  \begin{subfigure}[b]{0.136\linewidth}
    \includegraphics[width=\linewidth]{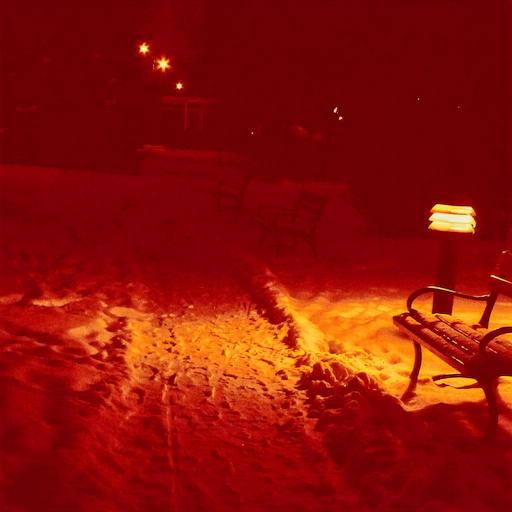}
  \end{subfigure}
  \sixcolgap
  \begin{subfigure}[b]{0.136\linewidth}
    \includegraphics[width=\linewidth]{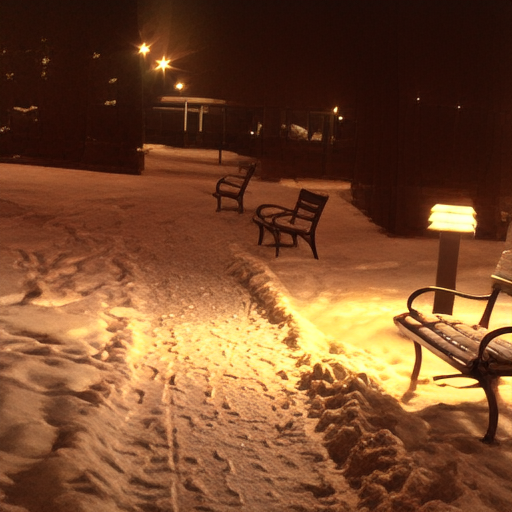}
  \end{subfigure}}

  \par\smallskip

  {\footnotesize\makebox[\linewidth][c]{%
  \begin{minipage}[t]{0.136\linewidth}
    \centering\textbf{Input}
  \end{minipage}
  \sixcolgap
  \begin{minipage}[t]{0.136\linewidth}
    \centering\textbf{Zero-DCE}
  \end{minipage}
  \sixcolgap
  \begin{minipage}[t]{0.136\linewidth}
    \centering\textbf{Retinexformer}
  \end{minipage}
  \sixcolgap
  \begin{minipage}[t]{0.136\linewidth}
    \centering\textbf{Reti-Diff}
  \end{minipage}
  \sixcolgap
  \begin{minipage}[t]{0.136\linewidth}
    \centering\textbf{DarkIR}
  \end{minipage}
  \sixcolgap
  \begin{minipage}[t]{0.136\linewidth}
    \centering\textbf{LUCID ($\beta = 1.05$)}
  \end{minipage}}}

  \caption{Visual comparison on the ExDark dataset. This comparison focuses on evaluating the enhancement performance on authentic night scenes.}
  \Description{A large comparison grid on ExDark scenes showing input images and outputs from several low-light enhancement methods, with LUCID producing cleaner and better exposed results.}
  \label{fig:exdark}
\end{figure*}

\begin{figure*}[p]
  \centering
  \makebox[\linewidth][c]{%
  \begin{subfigure}[b]{0.136\linewidth}
    \includegraphics[width=\linewidth]{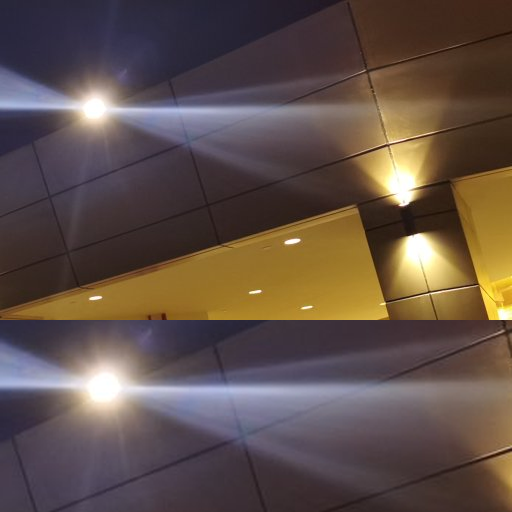}
  \end{subfigure}
  \sixcolgap
  \begin{subfigure}[b]{0.136\linewidth}
    \includegraphics[width=\linewidth]{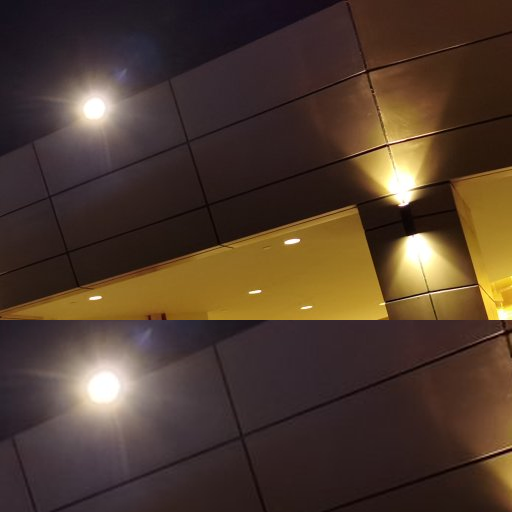}
  \end{subfigure}
  \sixcolgap
  \begin{subfigure}[b]{0.136\linewidth}
    \includegraphics[width=\linewidth]{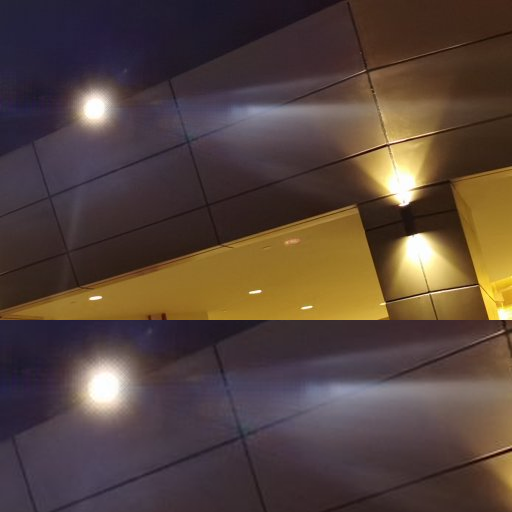}
  \end{subfigure}
  \sixcolgap
  \begin{subfigure}[b]{0.136\linewidth}
    \includegraphics[width=\linewidth]{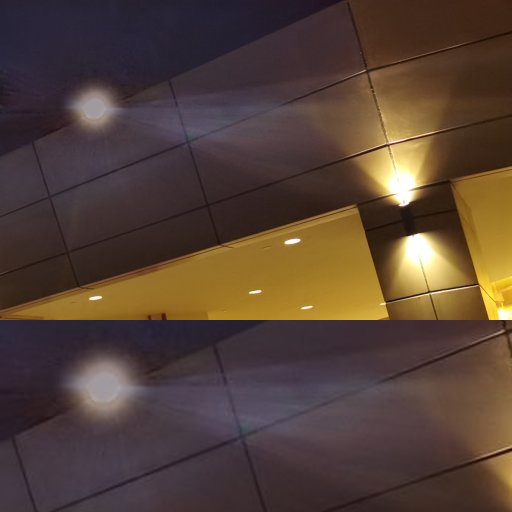}
  \end{subfigure}
  \sixcolgap
  \begin{subfigure}[b]{0.136\linewidth}
    \includegraphics[width=\linewidth]{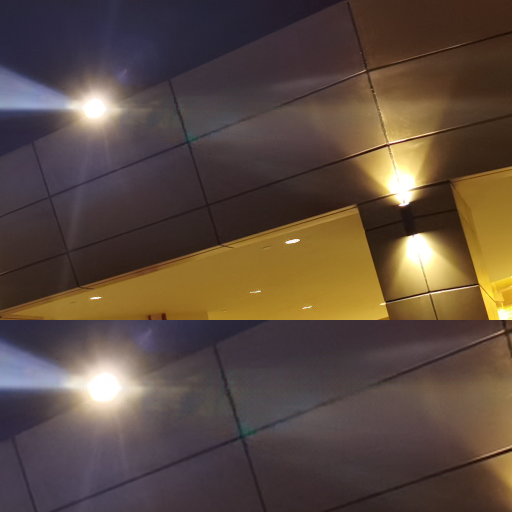}
  \end{subfigure}
  \sixcolgap
  \begin{subfigure}[b]{0.136\linewidth}
    \includegraphics[width=\linewidth]{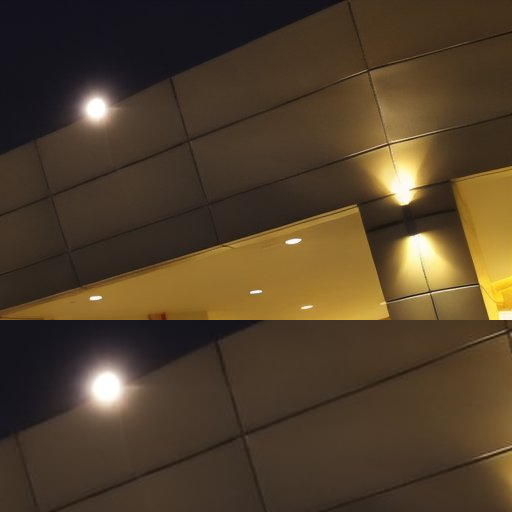}
  \end{subfigure}}

  \par\smallskip

  \makebox[\linewidth][c]{%
  \begin{subfigure}[b]{0.136\linewidth}
    \includegraphics[width=\linewidth]{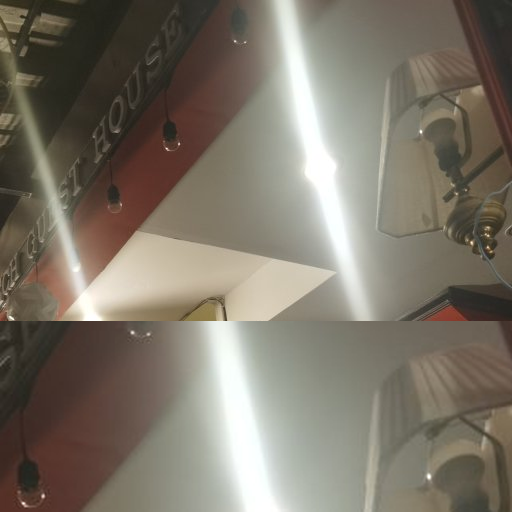}
  \end{subfigure}
  \sixcolgap
  \begin{subfigure}[b]{0.136\linewidth}
    \includegraphics[width=\linewidth]{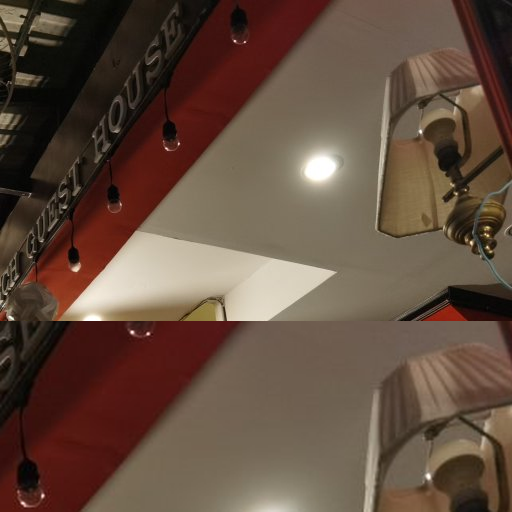}
  \end{subfigure}
  \sixcolgap
  \begin{subfigure}[b]{0.136\linewidth}
    \includegraphics[width=\linewidth]{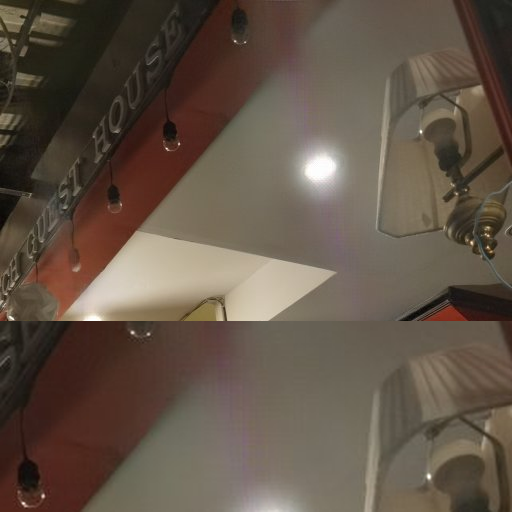}
  \end{subfigure}
  \sixcolgap
  \begin{subfigure}[b]{0.136\linewidth}
    \includegraphics[width=\linewidth]{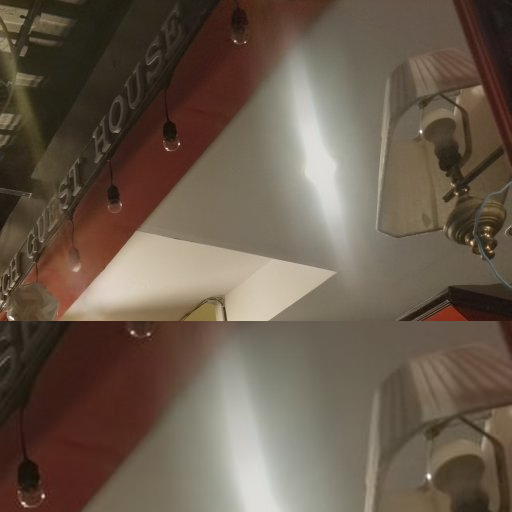}
  \end{subfigure}
  \sixcolgap
  \begin{subfigure}[b]{0.136\linewidth}
    \includegraphics[width=\linewidth]{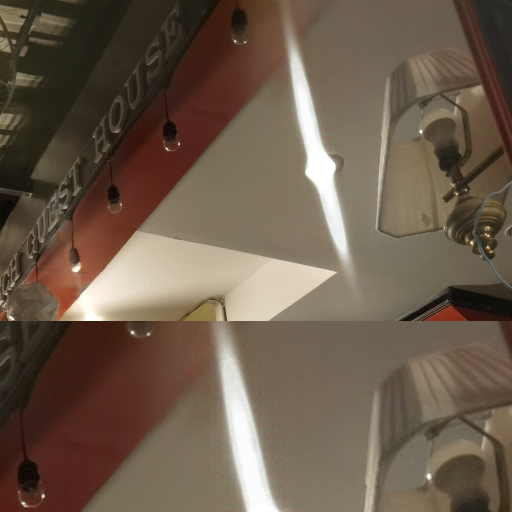}
  \end{subfigure}
  \sixcolgap
  \begin{subfigure}[b]{0.136\linewidth}
    \includegraphics[width=\linewidth]{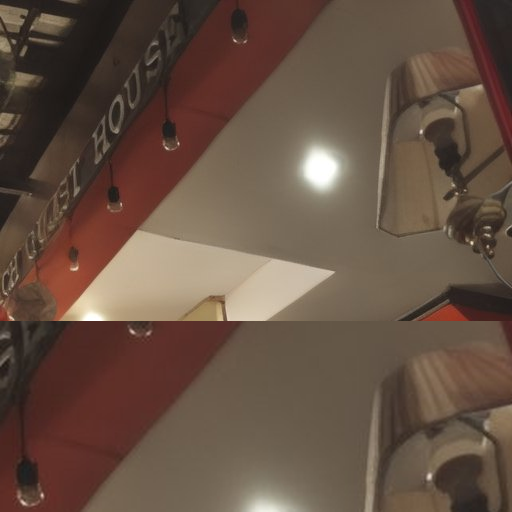}
  \end{subfigure}}

  \par\smallskip

  \makebox[\linewidth][c]{%
  \begin{subfigure}[b]{0.136\linewidth}
    \includegraphics[width=\linewidth]{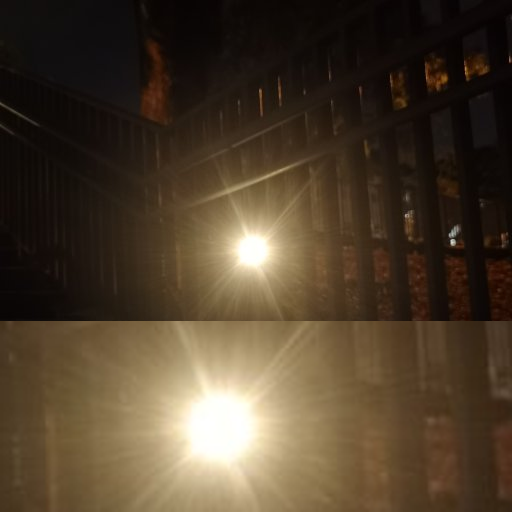}
  \end{subfigure}
  \sixcolgap
  \begin{subfigure}[b]{0.136\linewidth}
    \includegraphics[width=\linewidth]{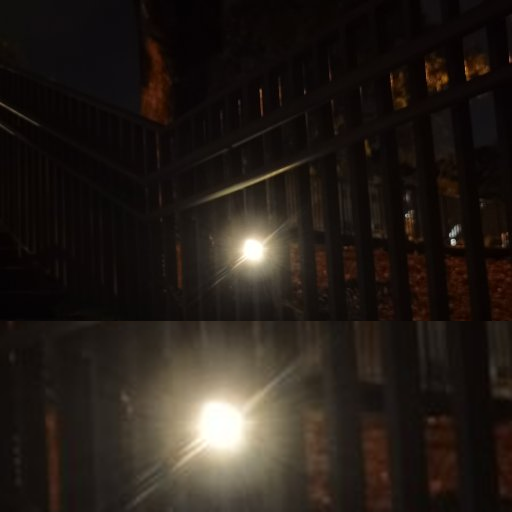}
  \end{subfigure}
  \sixcolgap
  \begin{subfigure}[b]{0.136\linewidth}
    \includegraphics[width=\linewidth]{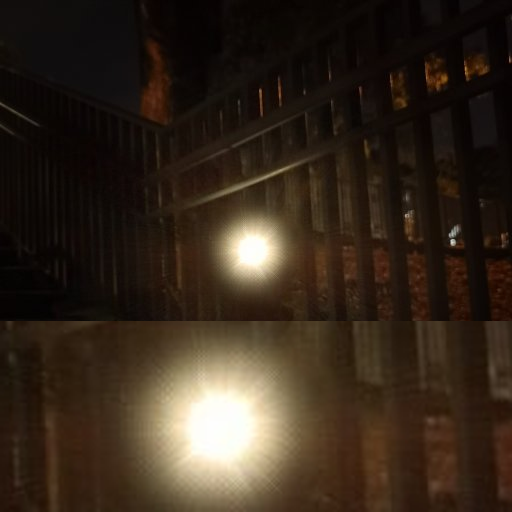}
  \end{subfigure}
  \sixcolgap
  \begin{subfigure}[b]{0.136\linewidth}
    \includegraphics[width=\linewidth]{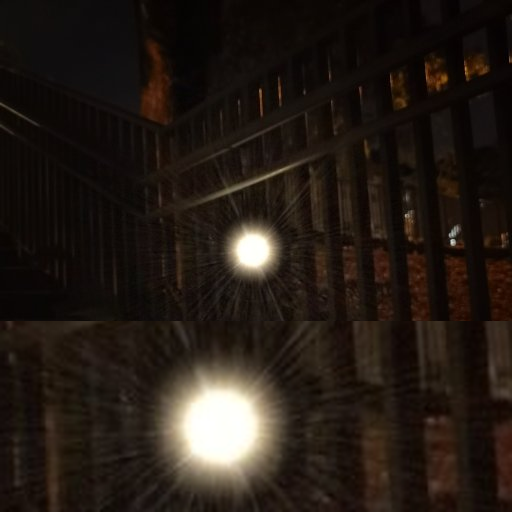}
  \end{subfigure}
  \sixcolgap
  \begin{subfigure}[b]{0.136\linewidth}
    \includegraphics[width=\linewidth]{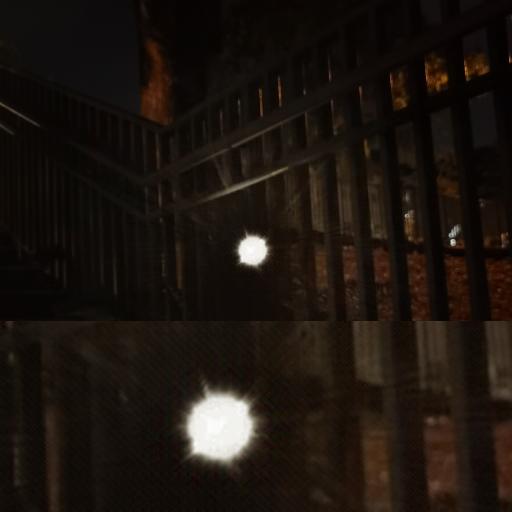}
  \end{subfigure}
  \sixcolgap
  \begin{subfigure}[b]{0.136\linewidth}
    \includegraphics[width=\linewidth]{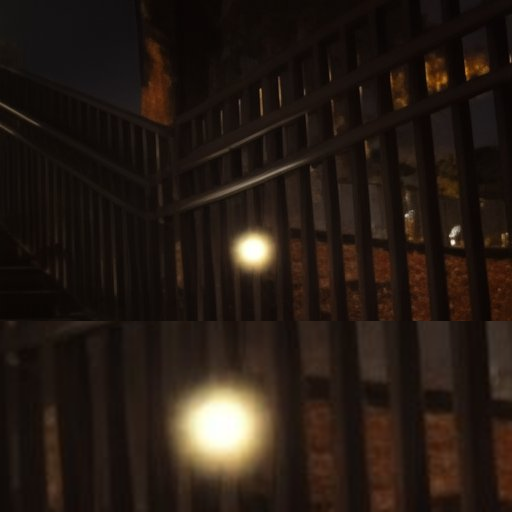}
  \end{subfigure}}

  \par\smallskip

  {\footnotesize\makebox[\linewidth][c]{%
  \begin{minipage}[t]{0.136\linewidth}
    \centering\textbf{Input}
  \end{minipage}
  \sixcolgap
  \begin{minipage}[t]{0.136\linewidth}
    \centering\textbf{GT}
  \end{minipage}
  \sixcolgap
  \begin{minipage}[t]{0.136\linewidth}
    \centering\textbf{Flare7K}
  \end{minipage}
  \sixcolgap
  \begin{minipage}[t]{0.136\linewidth}
    \centering\textbf{MFDNet}
  \end{minipage}
  \sixcolgap
  \begin{minipage}[t]{0.136\linewidth}
    \centering\textbf{Zhou \textit{et al.}}
  \end{minipage}
  \sixcolgap
  \begin{minipage}[t]{0.136\linewidth}
    \centering\textbf{LUCID ($\beta = 0.5$)}
  \end{minipage}}}

  \caption{Visual comparison on the Flare7K dataset. The comparison centers on the effectiveness of flare mitigation and preservation of light sources.}
  \Description{A large comparison grid on Flare7K scenes showing flare removal quality, light-source preservation, and qualitative differences between baselines and LUCID.}
  \label{fig:flare_results}
\end{figure*}

\begin{figure*}[p]
  \centering
  \makebox[\linewidth][c]{%
  \begin{subfigure}[b]{0.150\linewidth}
    \includegraphics[width=\linewidth]{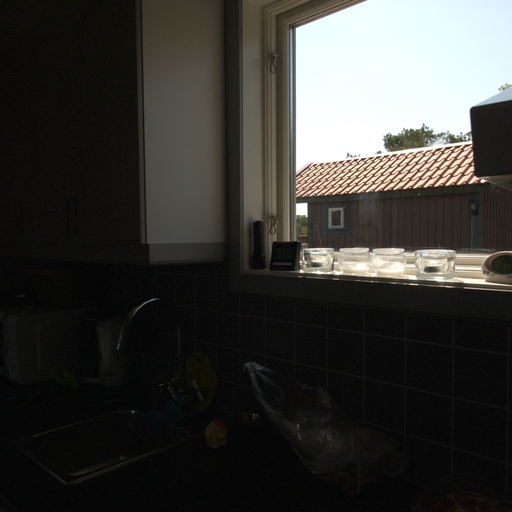}
  \end{subfigure}
  \fivecolgap
  \begin{subfigure}[b]{0.150\linewidth}
    \includegraphics[width=\linewidth]{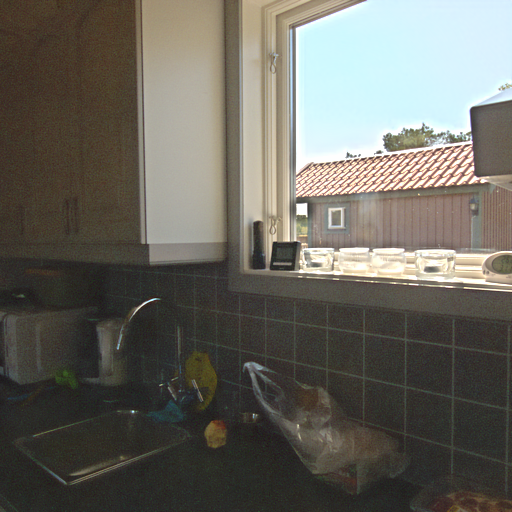}
  \end{subfigure}
  \fivecolgap
  \begin{subfigure}[b]{0.150\linewidth}
    \includegraphics[width=\linewidth]{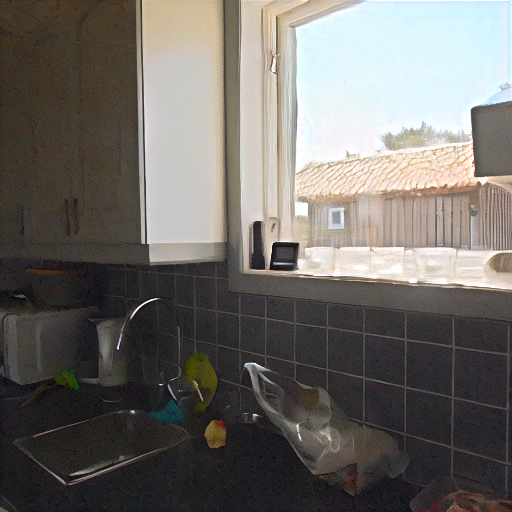}
  \end{subfigure}
  \fivecolgap
  \begin{subfigure}[b]{0.150\linewidth}
    \includegraphics[width=\linewidth]{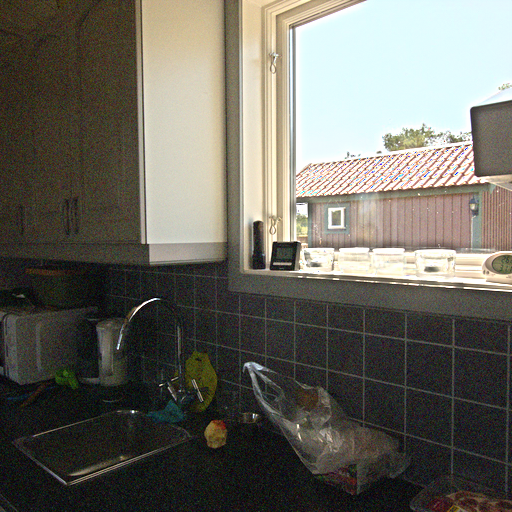}
  \end{subfigure}
  \fivecolgap
  \begin{subfigure}[b]{0.150\linewidth}
    \includegraphics[width=\linewidth]{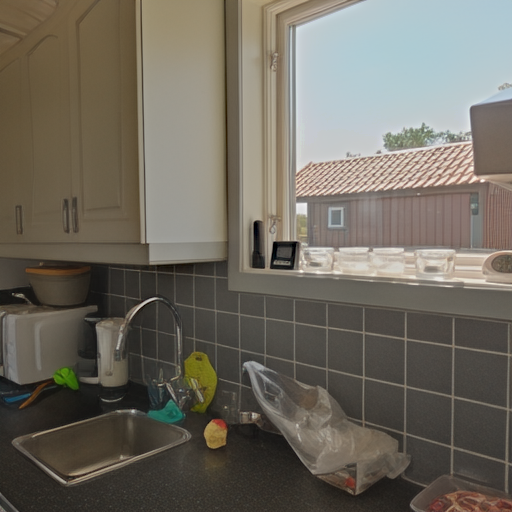}
  \end{subfigure}}

  \par\smallskip

  \makebox[\linewidth][c]{%
  \begin{subfigure}[b]{0.150\linewidth}
    \includegraphics[width=\linewidth]{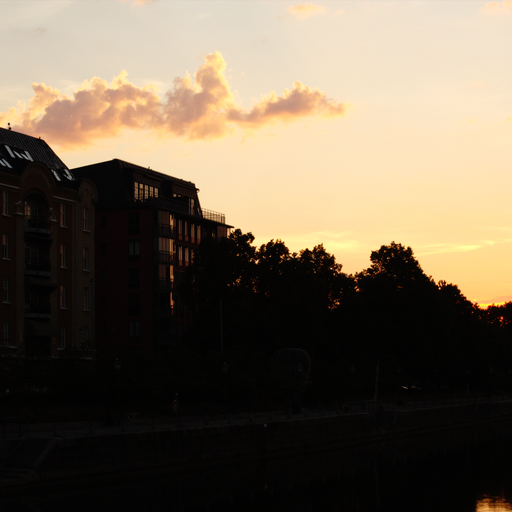}
  \end{subfigure}
  \fivecolgap
  \begin{subfigure}[b]{0.150\linewidth}
    \includegraphics[width=\linewidth]{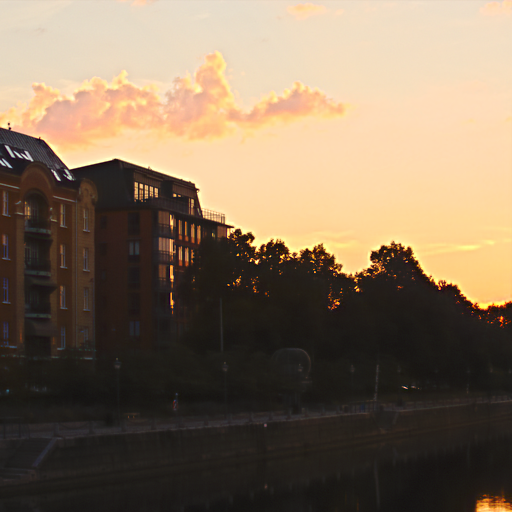}
  \end{subfigure}
  \fivecolgap
  \begin{subfigure}[b]{0.150\linewidth}
    \includegraphics[width=\linewidth]{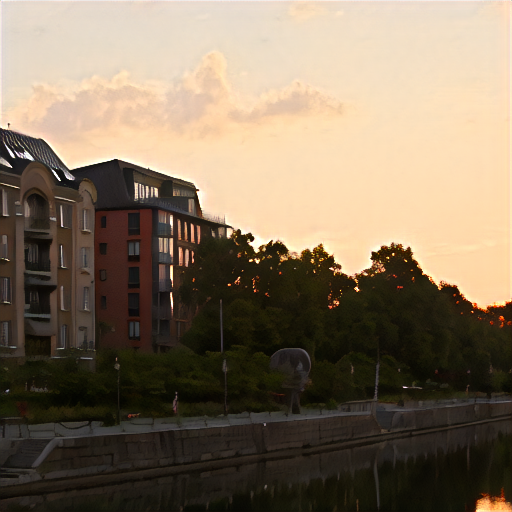}
  \end{subfigure}
  \fivecolgap
  \begin{subfigure}[b]{0.150\linewidth}
    \includegraphics[width=\linewidth]{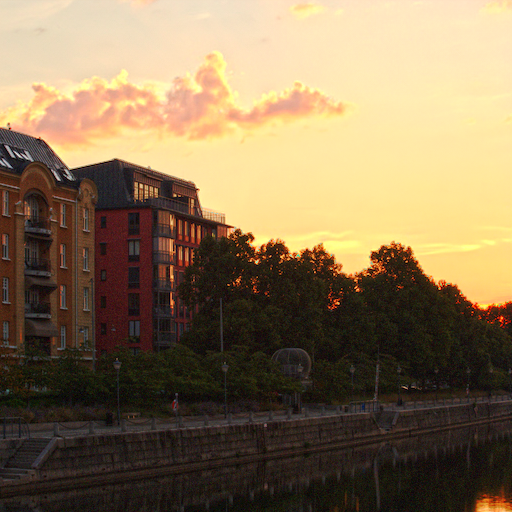}
  \end{subfigure}
  \fivecolgap
  \begin{subfigure}[b]{0.150\linewidth}
    \includegraphics[width=\linewidth]{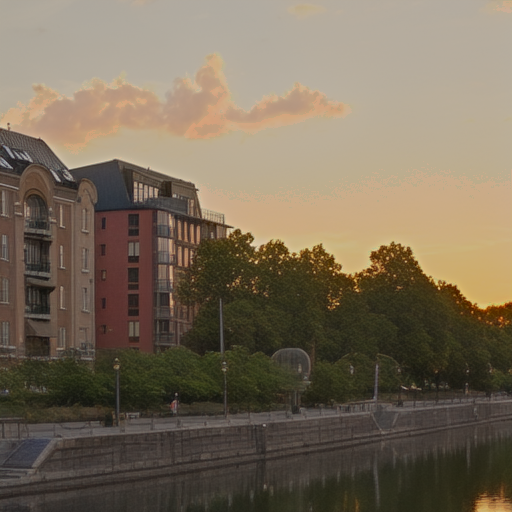}
  \end{subfigure}}

  \par\smallskip

  \makebox[\linewidth][c]{%
  \begin{subfigure}[b]{0.150\linewidth}
    \includegraphics[width=\linewidth]{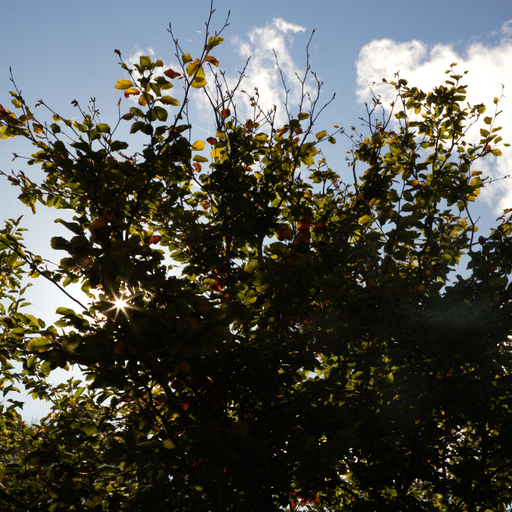}
  \end{subfigure}
  \fivecolgap
  \begin{subfigure}[b]{0.150\linewidth}
    \includegraphics[width=\linewidth]{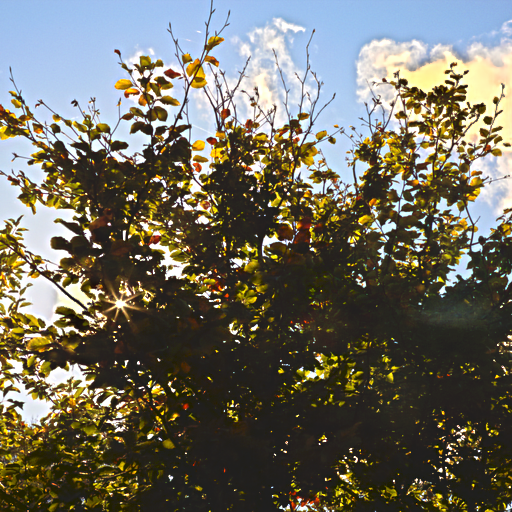}
  \end{subfigure}
  \fivecolgap
  \begin{subfigure}[b]{0.150\linewidth}
    \includegraphics[width=\linewidth]{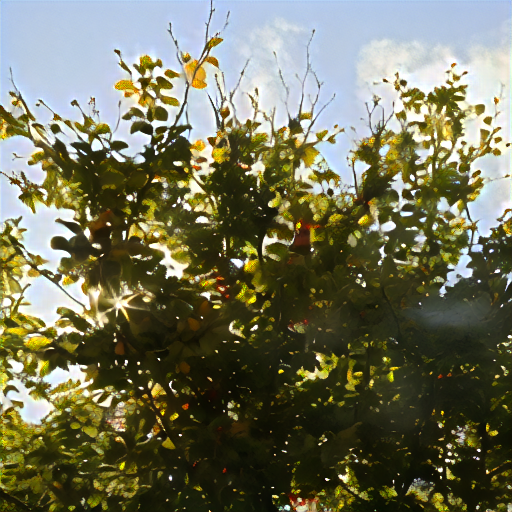}
  \end{subfigure}
  \fivecolgap
  \begin{subfigure}[b]{0.150\linewidth}
    \includegraphics[width=\linewidth]{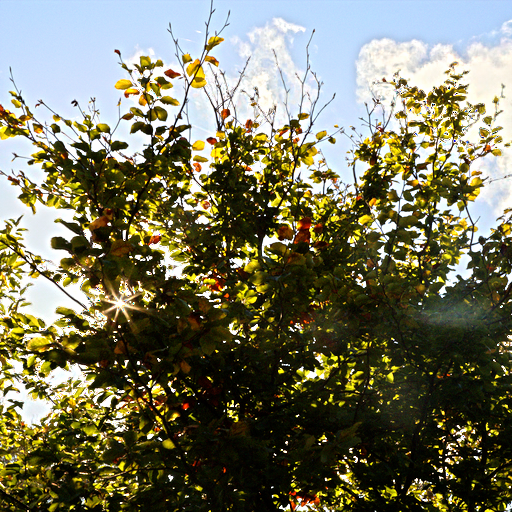}
  \end{subfigure}
  \fivecolgap
  \begin{subfigure}[b]{0.150\linewidth}
    \includegraphics[width=\linewidth]{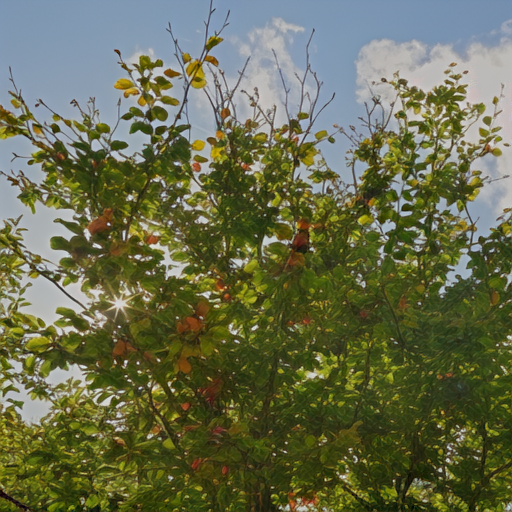}
  \end{subfigure}}

  \par\smallskip

  {\footnotesize\makebox[\linewidth][c]{%
  \begin{minipage}[t]{0.150\linewidth}
    \centering\textbf{Input}
  \end{minipage}
  \fivecolgap
  \begin{minipage}[t]{0.150\linewidth}
    \centering\textbf{IntrinsicHDR}
  \end{minipage}
  \fivecolgap
  \begin{minipage}[t]{0.150\linewidth}
    \centering\textbf{LEDiff}
  \end{minipage}
  \fivecolgap
  \begin{minipage}[t]{0.150\linewidth}
    \centering\textbf{GasLight}
  \end{minipage}
  \fivecolgap
  \begin{minipage}[t]{0.150\linewidth}
    \centering\textbf{LUCID (ours)}
  \end{minipage}}}

  \caption{Visual comparison on the SiHDR dataset. The comparison centers on HDR reconstruction.}
  \Description{A large comparison grid on SiHDR examples showing HDR reconstructions from baseline methods and LUCID, emphasizing highlight recovery and shadow detail.}
  \label{fig:hdr_results}
\end{figure*}

\begin{figure*}[p]
  \centering
  \includegraphics[width=\linewidth]{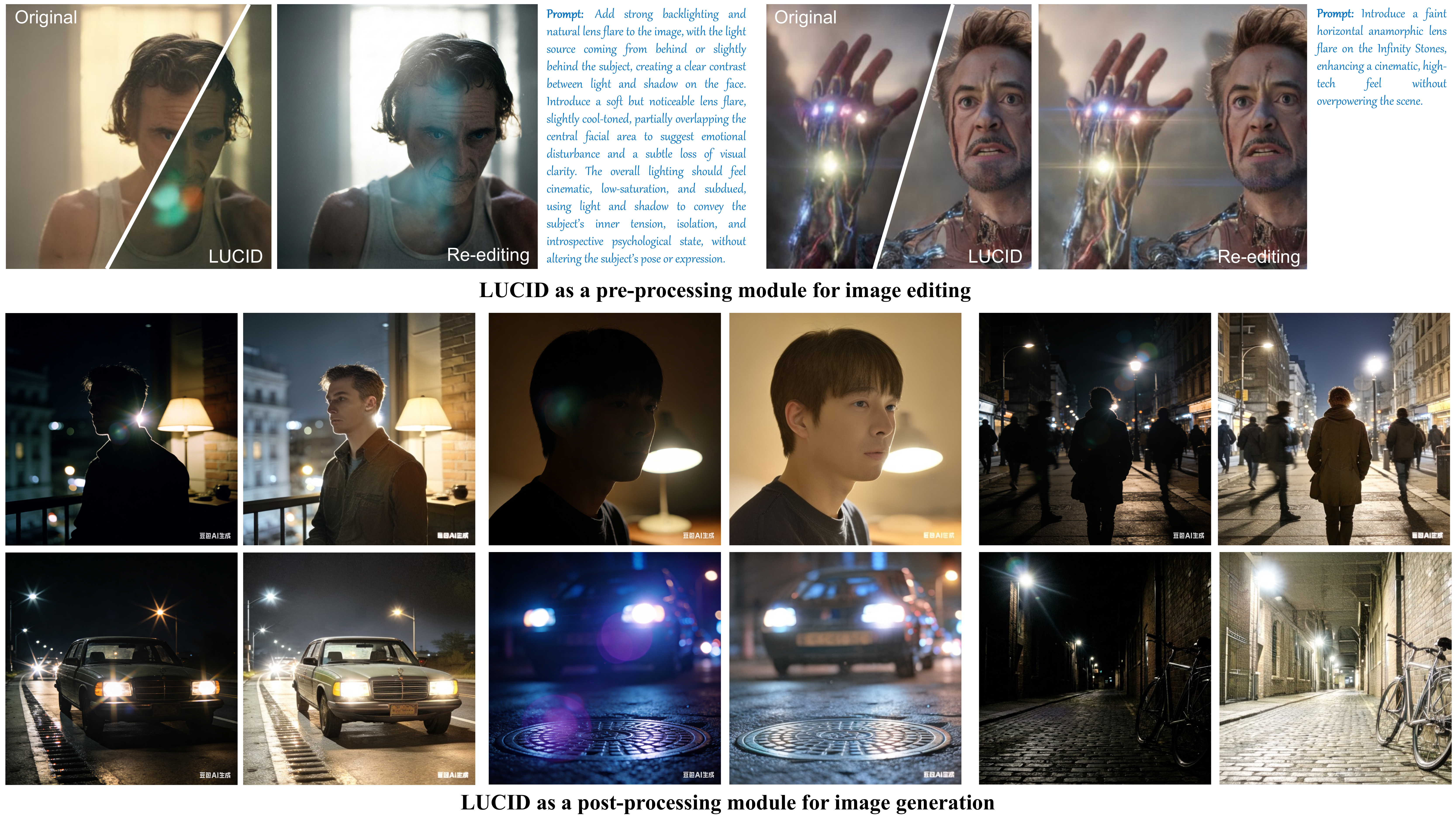}
  \caption{LUCID enables creative image editing. Top: LUCID suppresses existing lens flare, effectively decoupling flare artifacts from the underlying image and preparing the result for subsequent editing (using Nano-Banana Pro). Selected film stills are sourced from ShotDeck for academic study. Bottom: combined with AI image generators (Doubao AI and Nano-Banana Pro), LUCID serves as a post-processing module for flexible adjustment of exposure and lens flare.}
  \Description{A creative-editing figure showing LUCID used before and after commercial image generation to remove flare, edit scenes, and reintroduce controlled nighttime lighting effects.}
  \label{fig:creative_results}
\end{figure*}

\begin{figure*}[t]
  \centering
  \includegraphics[width=\linewidth]{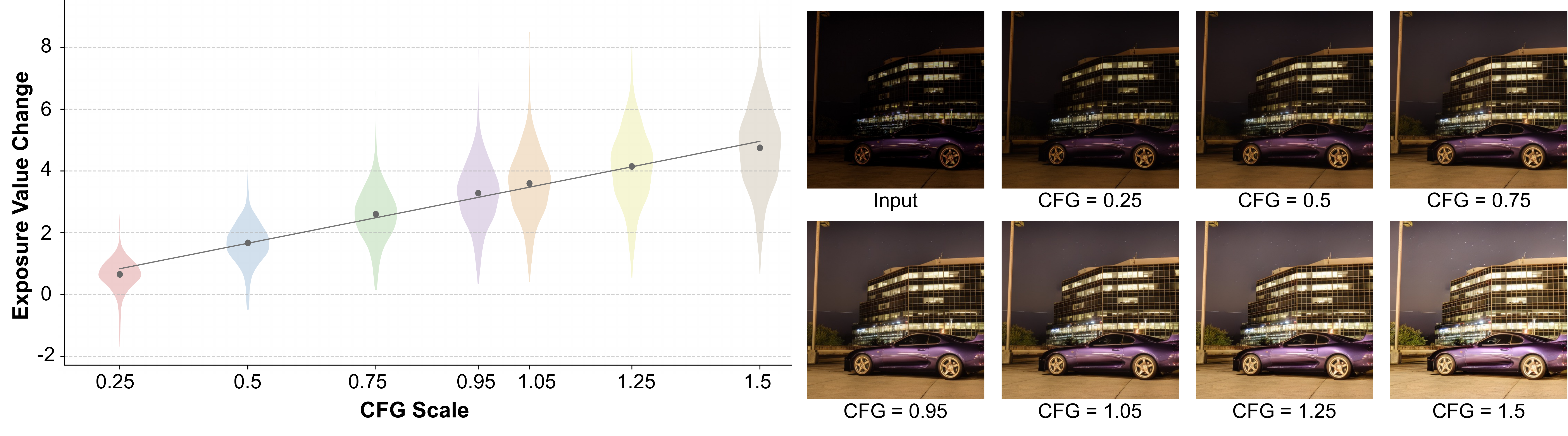}
  \caption{Relationship between CFG scales (\(\beta\)) and exposure value changes ($\Delta$EV). Left: Violin plots demonstrate a predictable linear response under a standard sRGB gamma mapping ($\gamma=2.2$). Right: Visual examples verify that adjusting \(\beta\) achieves perceptually smooth exposure ramping.}
  \Description{A figure with violin plots and sample images showing that increasing CFG scale produces a smooth, approximately linear increase in exposure.}
  \label{fig:CFG_EV}
\end{figure*}

\subsection{Results}

\subsubsection{Performance on Holistic Nighttime Restoration.} Fig.~\ref{fig:exdark} validates the robustness of LUCID across four diverse nighttime regimes (top to bottom): extreme photon starvation, high-contrast dynamic range, strong backlighting with veiling glare, and severe illuminant color casts. Baseline methods exhibit characteristic failures: they linearly amplify sensor noise in signal-starved regions (Row 1), over-expose localized headlights (Row 2), struggle to penetrate optical artifacts (Row 3), or blindly enhance the dominant hue, causing unnatural chromatic distortion. In contrast, LUCID demonstrates superior semantic consistency. It effectively suppresses noise in ultra-dark limits and compresses dynamic range to recover highlight textures. Simultaneously, it rectifies color deviations and restores clarity in washed-out backlit scenarios. This confirms its versatility as a robust solution for uncontrolled real-world nighttime imaging.

Quantitative results based on no-reference perceptual metrics are reported in Tab.~\ref{tab:iqa_comparison}. These results statistically validate that our method not only enhances visibility but also aligns better with human perceptual preferences, demonstrating superior generalization to diverse authentic nighttime imagery.

\begin{table}[t]
  \centering
  \setlength{\tabcolsep}{4pt}
  \caption{Quantitative comparison on IQA metrics. All metrics are reference-free, higher is better ($\uparrow$). Best results are \textbf{bolded}. We select a fixed $\beta = 1.05$ based on a balance between CLIPIQA and MANIQA.}
  \begin{tabular}{lccccc}
    \toprule
    Method & CLIPIQA & MANIQA & MUSIQ & LIQE & NIMA \\
    \midrule
    ExDark & 0.4281 & 0.2939 & 50.99 & 2.267 & 5.227 \\
    Zero-DCE & 0.4243 & 0.2899 & 50.98 & 2.078 & 4.999 \\
    Retinexformer & 0.3749 & 0.2588 & 52.26 & 2.058 & 5.121 \\
    Jin \textit{et al.} & 0.3664 & 0.2585 & 51.89 & 2.087 & 5.078 \\
    Reti-Diff & 0.4159 & 0.2867 & 50.91 & 1.967 & 5.083 \\
    DarkIR & 0.4107 & 0.3078 & 52.03 & 2.076 & 5.046 \\
    \midrule
    \textbf{Ours ($\beta = 1.05$)} & \textbf{0.4774} & \textbf{0.3264} & \textbf{61.45} & \textbf{3.019} & \textbf{5.390} \\
    \bottomrule
  \end{tabular}
\label{tab:iqa_comparison}
\end{table}

\subsubsection{Flare Mitigation.} To isolate and evaluate flare mitigation performance, we conduct experiments on the Flare7K dataset. Crucially, the comparative baselines are specialized strictly for artifact removal, lacking the low-light enhancement capabilities integral to our method. Since the dataset is constructed by introducing physical smudges, the ground-truth (GT) may inherently retain slight residual flares from the lens optics. As shown in Fig.~\ref{fig:flare_results}, baseline methods struggle to achieve a balance between removal and preservation. They often exhibit incomplete removal, leaving residual streaks and hazy artifacts (1st and 2nd rows). Conversely, some methods tend to over-subtract the signal around light sources, resulting in unnatural sharp boundaries (e.g., Zhou et al. in the 3rd row). In contrast, LUCID successfully disentangles and removes the scattering artifacts. The proposed method leverages generative priors to synthesize reliable background textures while reconstructing the natural optical fall-off of light sources. The resulting images are visually clean and physically plausible, occasionally exceeding the quality of GT.

\begin{figure}[h]
  \centering
  \includegraphics[width=\linewidth]{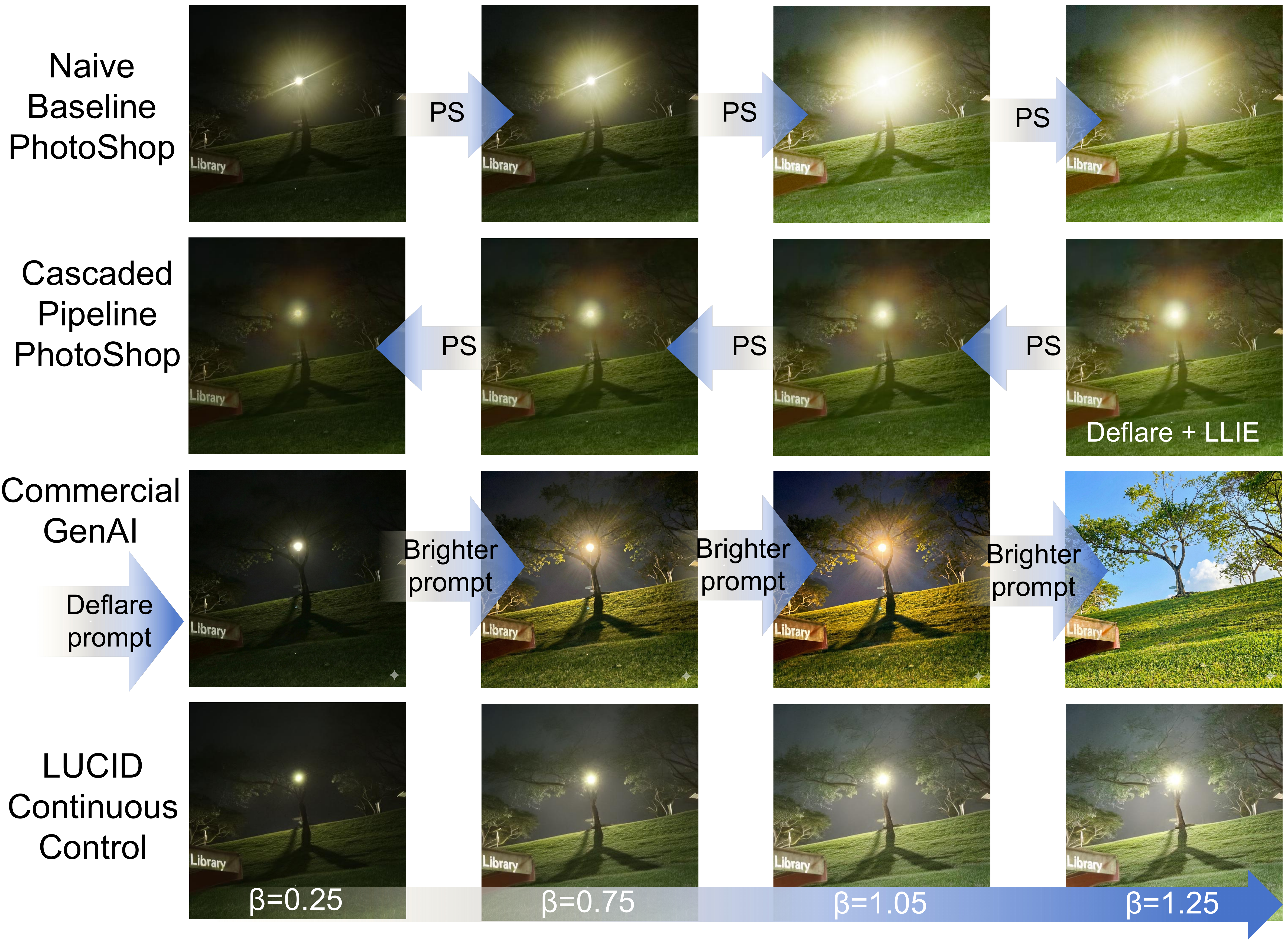}
  \caption{Visual comparison of controllability across different workflows.}
  \Description{A comparison figure showing that LUCID provides smoother and more faithful exposure control than naive editing, cascaded restoration, or commercial generative editing.}
  \label{fig:Comp}
\end{figure}

\begin{figure}[t]
  \centering
  \begin{subfigure}{0.32\linewidth}
    \includegraphics[width=\linewidth]{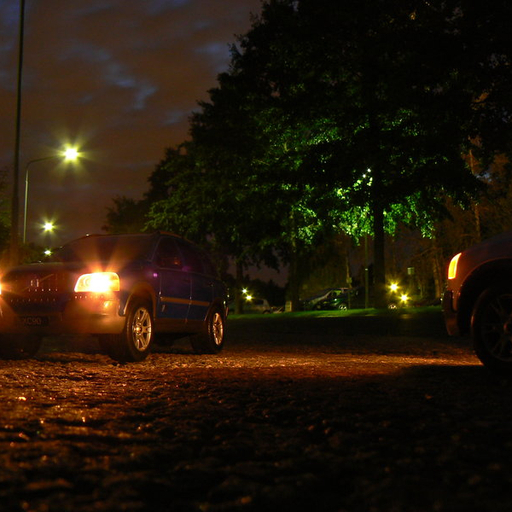}
  \end{subfigure}
  \hfill
  \begin{subfigure}{0.32\linewidth}
    \includegraphics[width=\linewidth]{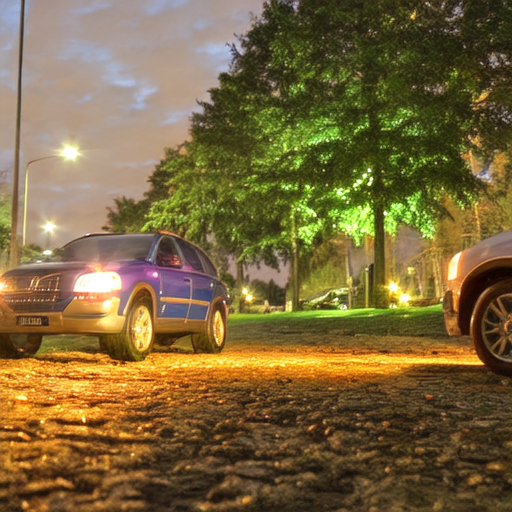}
  \end{subfigure}
  \hfill
  \begin{subfigure}{0.32\linewidth}
    \includegraphics[width=\linewidth]{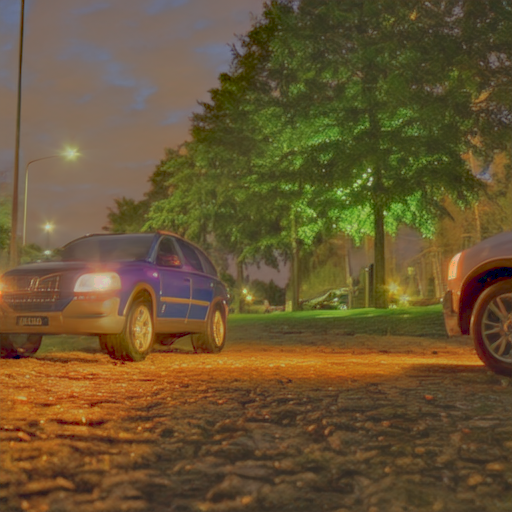}
  \end{subfigure}

  \medskip

  \begin{subfigure}{0.32\linewidth}
    \includegraphics[width=\linewidth]{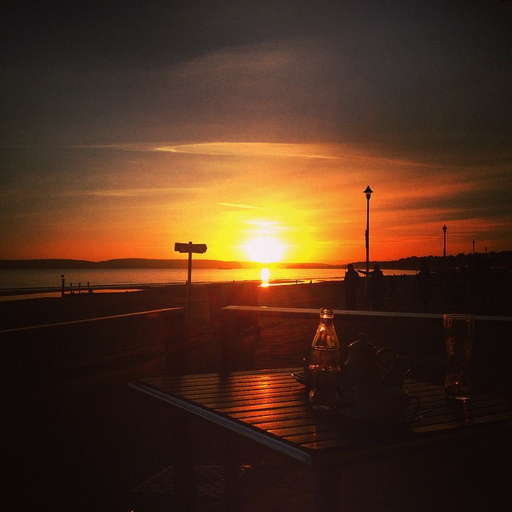}
    \caption{Input}
  \end{subfigure}
  \hfill
  \begin{subfigure}{0.32\linewidth}
    \includegraphics[width=\linewidth]{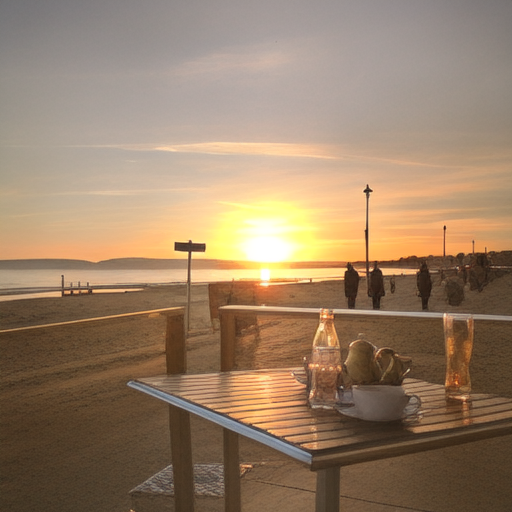}
    \caption{$\beta = 1.05$}
  \end{subfigure}
  \hfill
  \begin{subfigure}{0.32\linewidth}
    \includegraphics[width=\linewidth]{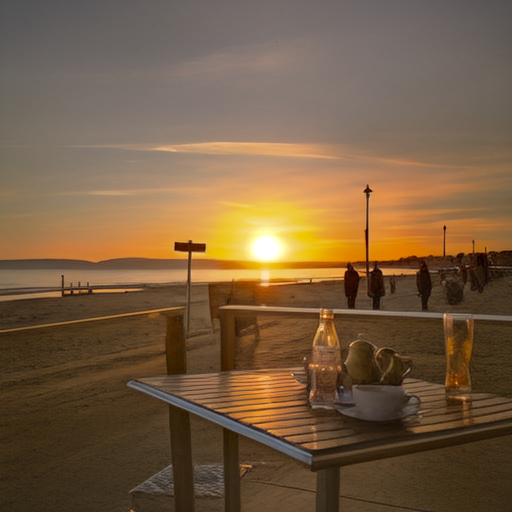}
    \caption{HDR}
  \end{subfigure}
  \caption{Dual restoration aesthetics. LUCID yields both perceptual realism at $\beta=1.05$ and an alternative HDR aesthetic, maximizing detail visibility across shadows and highlights.}
  \Description{A six-image figure comparing the input, a perceptually realistic restoration, and an HDR-style result to illustrate two aesthetic outputs from the same scene.}
  \label{fig:dual}
\end{figure}

\subsubsection{Continuous Control.} A core contribution of LUCID is empowering photographers with precise, continuous control over the restoration process. To validate the predictability of our control mechanism, we first analyze the statistical relationship between the CFG scale (\(\beta\)) and the relative exposure value change (\(\Delta\text{EV}\)). As visualized in Fig.~\ref{fig:CFG_EV}, the mean exposure value increment exhibits a monotonic and quasi-linear response to the guidance scale, allowing users to intuitively dial in the desired brightness. The visual examples visualize the continuous modulation effects. Spanning $\beta \in [0.25, 1.5]$, LUCID maintains robust structural consistency across all intervals while rendering smooth, natural transitions in illuminance. This progression effectively mimics the physical behavior of gradually intensifying a dimmer-controlled light source. We contrast our controllable paradigm against three alternatives (Fig.~\ref{fig:Comp}).

\paragraph{Naive Baseline}: Global exposure adjustment is physically flawed; it amplifies sensor noise and expands the radius of veiling glare, severely washing out scene contrast.

\paragraph{Cascaded Pipeline}: We sequentially chain a deflare network~\cite{dai2022flare7k} with a low-light enhancer~\cite{Feijoo_2025_CVPRdarkir}, followed by manual dimming (PhotoShop). This suffers from compound error accumulation, where residual artifacts missed by the deflare stage are aggressively amplified by the enhancer into unnatural residues.

\paragraph{Commercial GenAI} (Nano-Banana Pro): Despite impressive resolution, relying solely on text prompts proves insufficient: such control is too coarse for precise illuminance tuning and prone to semantic drift (e.g., hallucinating daylight structures).

Ultimately, LUCID transcends these limitations by reconciling generative fidelity with user intent. It provides a transparent, controllable framework that eliminates optical degradations. This capability offers a level of interaction and consistency that is typically absent in unconstrained, purely text-guided generative architectures.

\subsubsection{Single-Image HDR Reconstruction.} By fusing ``virtual exposure brackets'' synthesized via CFG control, LUCID extends naturally to HDR reconstruction. Fig.~\ref{fig:hdr_results} compares our results against specialized HDR methods. Baselines exhibit distinct failures: IntrinsicHDR leaves deep shadows crushed (Row 1); LEDiff and GasLight suffer from generative hallucinations (e.g., distorted roof tiles in Row 1), unnatural color shifts (e.g., the artificial orange cast in Row 2), and erroneously amplify lens flare (Row 3). In contrast, LUCID achieves superior photorealism, maintaining structural fidelity and natural color balance across all exposure levels. This introduces a flexible aesthetic dimension, enabling idealized scene rendering that transcends the imperfections of raw physical capture (Fig.~\ref{fig:dual}).

\subsubsection{Downstream Applications.} LUCID serves as a plug-and-play module for diverse workflows (Fig.~\ref{fig:creative_results}). As a pre-processor (top), it suppresses pre-existing flares to provide a clean, artifact-free input, ensuring subsequent prompt-driven editing remains free from original lighting interference. As a post-processor (bottom), it introduces post-hoc tunability to generative models, enabling flexible adjustment of exposure and flare on otherwise static AI-generated outputs within downstream creative pipelines.

\section{Conclusion and Limitation}

In this work, we introduce LUCID to address the entangled challenges of flare and low-light in nighttime photography. Unlike fixed-solution baselines, our framework enables continuous luminance modulation and precise exposure control. Extensive experiments demonstrate the superior generalization of LUCID in rendering night scenes with clear visibility and pristine aesthetics.

Admittedly, hallucination remains an unavoidable challenge for current generative models. However, in extremely challenging cases such as severe darkness, LUCID prioritizes conservative and controllable reconstruction over aggressively hallucinating ungrounded details. We provide further discussion and visual examples of this restrained generative behavior in the supplementary material.
We believe this paradigm of controllable generation paves the way for future intelligent computational photography tools.

\begin{acks}
This work is supported by the National Natural Science Foundation of China (NSFC) under Grants 62271283 and 62571319, and the Shanghai Jiao Tong University Medical and Engineering Cross Research Fund under Grant YG2026QNB50.
\end{acks}

\bibliographystyle{ACM-Reference-Format}
\bibliography{reference}

\input{}

\end{document}